%% file: main.tex
\documentclass[runningheads]{llncs}

\usepackage[mobile]{eccv}

\usepackage{multirow}
\usepackage[dvipsnames, svgnames, x11names, table, xcdraw]{xcolor}
\usepackage{microtype}
\usepackage{wrapfig}
\usepackage{eccvabbrv}

\usepackage{graphicx}
\usepackage{booktabs}

\usepackage[accsupp]{axessibility}  
\usepackage[breakable]{tcolorbox}
\renewcommand{\thefootnote}{\fnsymbol{footnote}}
\usepackage{hyperref}

\usepackage{orcidlink}

\begin{document}

\title{Concept-to-Pixel: Prompt-Free Universal Medical Image Segmentation} 



\renewcommand{\thefootnote}{}
\author{Haoyun Chen\inst{1,2,3\dagger} \and
Fenghe Tang\inst{1,2,3\dagger}\orcidlink{0009-0009-6193-4855} \and
Wenxin Ma\inst{1,2,3}\orcidlink{0009-0008-0967-2344} \and
Shaohua Kevin Zhou\inst{1,2,3,4}\thanks{Corresponding author $\dagger$ \text{Equal Contribution}}\orcidlink{0000-0002-6881-4444}}

\authorrunning{Haoyun Chen, Fenghe Tang, Wenxin Ma, Shaohua Kevin Zhou.}

\institute{School of Biomedical Engineering, Division of Life Sciences and Medicine, University of Science and Technology of China (USTC), Hefei, Anhui 230026, China \and
Center for Medical Imaging, Robotics, Analytic Computing \& Learning (MIRACLE), Suzhou Institute for Advanced Research, USTC, Suzhou, Jiangsu 215123, China\\
\and
Jiangsu Provincial Key Laboratory of Multimodal Digital Twin Technology, Suzhou Jiangsu, 215123, China\\
\and
State Key Laboratory of Precision and Intelligent Chemistry, USTC, Hefei Anhui 230026, China \\
\email{\{chenhaoyun,fhtan9,wxma\}@mail.ustc.edu.cn}, \email{\{skevinzhou\}@ustc.edu.cn}
}

\maketitle

\vspace{-10mm}
\begin{abstract}
Universal medical image segmentation seeks to use a single foundational model to handle diverse tasks across multiple imaging modalities. However, existing approaches often rely heavily on manual visual prompts or retrieved reference images, which limits their automation and robustness. In addition, naive joint training across modalities often fails to address large domain shifts. To address these limitations, we propose \textbf{Concept-to-Pixel (C2P)}, a novel prompt-free universal segmentation framework. C2P explicitly separates anatomical knowledge into two components: Geometric and Semantic representations. It leverages Multimodal Large Language Models (MLLMs) to distill abstract, high-level medical concepts into learnable Semantic Tokens and introduces explicitly supervised Geometric Tokens to enforce universal physical and structural constraints.
These disentangled tokens interact deeply with image features to generate input-specific dynamic kernels for precise mask prediction. Furthermore, we introduce a Geometry-Aware Inference Consensus mechanism, which utilizes the model's predicted geometric constraints to assess prediction reliability and suppress outliers. Extensive experiments and analysis on a unified benchmark comprising eight diverse datasets across seven modalities demonstrate the significant superiority of our jointly trained approach, compared to universe- or single-model approaches. Remarkably, our unified model demonstrates strong generalization, achieving impressive results not only on zero-shot tasks involving unseen cases but also in cross-modal transfers across similar tasks. Code is available at: \textcolor{magenta}{\url{https://github.com/Yundi218/Concept-to-Pixel}}.

\end{abstract}

\section{Introduction}
\label{sec:intro}

Medical image segmentation is an important task of computer-aided diagnosis~\cite{zhou2017deep,tang2025u,hyspark,mambamim,hiendmae,medgmae}, but the vast diversity of modalities and severe domain gaps in tissue contrast and noise distributions have historically confined the field to specialized, modality-specific solutions~\cite{isensee2021nnu,ronneberger2015u,ye2025u,liu2024swin,cmunet,cmunext,mobileuvit,llm4seg}. Beyond their computational cost and poor scalability, these ``one-model-one-task'' architectures fundamentally fail to exploit cross-modal anatomical knowledge for mutual performance gains. This has inspired a paradigm shift toward universal segmentation models~\cite{butoi2023universeg,rakic2024tyche,zhao2024spider,chang2025unified,Ma_2025_WACV}, which aim to unify diverse segmentation tasks across modalities within a single framework.

\begin{figure}[t] 
  \centering
  \includegraphics[width=\textwidth]{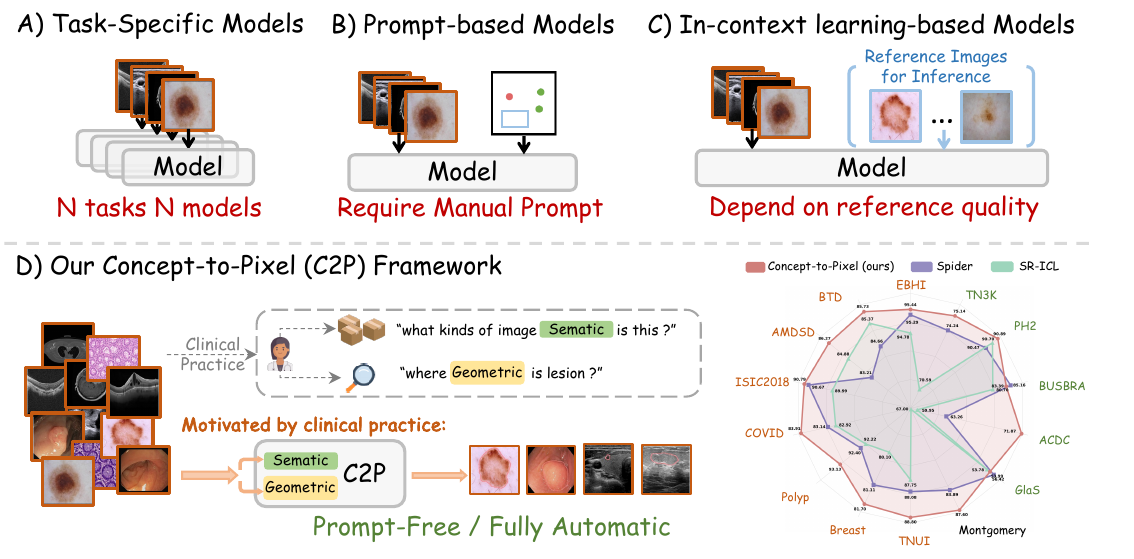} 
  \caption{\textbf{Comparison of existing medical image segmentation paradigms and our proposed Concept-to-Pixel framework.} Unlike existing paradigms that suffer from limited versatility \textbf{(A)}, limited manual prompts \textbf{(B)}, or reliance on reference quality \textbf{(C)}, our prompt-free model \textbf{(D)}, inspired by clinical diagnostic workflows, explicitly decouples modality semantics and physical geometry via \texttt{[SEM]} and \texttt{[GEO]}. The radar chart in \textbf{(D) Right} demonstrates that our zero-reference method (\textcolor[HTML]{D07F7A}{Concept-to-Pixel}) significantly outperforms SOTA models (\textcolor[HTML]{8379B6}{Spider} and \textcolor[HTML]{9DD5BC}{SR-ICL}), exhibiting impressive \textcolor[HTML]{C4570C}{in-domain performance} and \textcolor[HTML]{4E7E2D}{zero-shot generalization} even on completely unseen modalities (\eg, X-Ray).}
  \label{fig:teaser}
\end{figure}

However, building such a unified model that matches or exceeds specialized counterparts remains extremely difficult. As shown in~\cref{fig:teaser} (B,C), existing approaches fall into three categories: (1) \textbf{Prompt-based methods}~\cite{cheng2023sam,ma2024segment,slidesam,wang2024video} excel at interactive boundary delineation but rely heavily on manual visual prompts (points or boxes), lacking the semantic understanding necessary for fully automatic lesion and anatomy discovery. (2) \textbf{In-context learning (ICL) methods}~\cite{zhao2024spider,chang2025unified} perform segmentation by conditioning on reference image-mask pairs, but this inference-heavy paradigm suffers from high variance and depends critically on the quality and relevance of retrieved support examples, making it difficult to globally internalize medical knowledge. (3) \textbf{Naive joint training} on large-scale heterogeneous datasets frequently induces negative transfer: conflicting low-level visual cues from different input sources drive gradients in opposing directions~\cite{vandenhende2021multi}. For example, a dark hypoechoic region signals malignancy in ultrasound yet represents healthy fluid in MRI, forcing the network to implicitly reconcile irreconcilable pixel-level patterns and ultimately causing feature space collapse.

Previous works~\cite{claim5,claim6} reveal that the root cause of this collapse is that the network is forced to reconcile conflicts at the pixel level, without any structured intermediate representation. Structured clinical diagnosis, by contrast, naturally avoids such confusion by evaluating \emph{disentangled concepts}: a clinician assesses a lesion by implicitly decoupling its \emph{Geometry}, including modality-agnostic physical properties such as shape, location, and boundaries, from its \emph{Semantics}, which is modality-specific textural appearance and internal signal characteristics~\cite{claim1,claim2,claim3,claim4}. We illustrate the procedure in~\cref{fig:teaser} (D).

Motivated by this diagnostic principle, we propose \textbf{Concept-to-Pixel (C2P)}, a novel framework that fundamentally mitigates negative transfer by explicitly disentangling anatomical understanding into separate Geometric and Semantic representations.
Specifically, to enforce universal shape priors, we introduce learnable Geometric Tokens, marked as \texttt{[GEO]}, that are explicitly supervised to regress physical attributes (\eg, area, centroid), imposing a strict structural constraint largely ignored by conventional U-Nets. For semantic understanding, we leverage Multimodal Large Language Models (MLLMs) to generate and extract fine-grained concepts from clinical reports (\eg, ``irregular margin'', ``heterogeneous texture''). We distill these text concepts into Semantic Tokens marked as \texttt{[SEM]}.
Subsequently, \texttt{[GEO]} and \texttt{[SEM]} serve as the sample-aware priors, interacting with visual features through a Token-Guided Dynamic Head (TGDH) to generate sample-aware dynamic segmentation head~\cite{tian2020conditional,jia2016dynamic}.
Furthermore, we introduce an intrinsic Geometry-Aware Inference Consensus mechanism, utilizing the explicitly regressed geometric priors to self-evaluate prediction reliability across perturbed views and confidently suppress outliers. 
Together, these designs establish C2P as a principled and generalizable framework. Extensive experiments have proven that our method achieves state-of-the-art performance across eight datasets spanning seven modalities with a single universal model. Remarkably, our C2P demonstrates strong zero-shot generalization on five unseen datasets without relying on any references or prompts, and attains competitive performance even on entirely unseen modalities.

Our main contributions are summarized as follows:
\begin{itemize}
\item We propose \textbf{Concept-to-Pixel (C2P)}, a universal segmentation framework that decouples anatomical understanding into \texttt{[GEO]} (supervised by physical attributes) and \texttt{[SEM]} (distilled from MLLM-extracted clinical concepts), enabling modality-agnostic representations that mitigate negative transfer in heterogeneous multi-modal training.

\item While the learned tokens capture universal geometric and semantic representations, we design a \textbf{Token-Guided Dynamic Head (TGDH)} that conditions on per-sample tokens to generate instance-specific convolution kernels, enabling the model to adapt its segmentation behavior to each individual sample rather than applying a shared static head across all targets.

\item Extensive experiments on eight datasets across seven modalities show that C2P, as a \textit{single} universal model, achieves {88.22\%} average Dice, surpassing both state-of-the-art universal methods and the task-specialized nnU-Net V2. Additionally, our C2P demonstrates strong generalization ability without requiring additional prompts or references, achieving competitive results on unseen datasets and modalities.

\end{itemize}

\section{Method}
\label{sec:method}

\subsection{Overview}
\begin{figure}[t] 
  \centering
  \includegraphics[width=\textwidth]{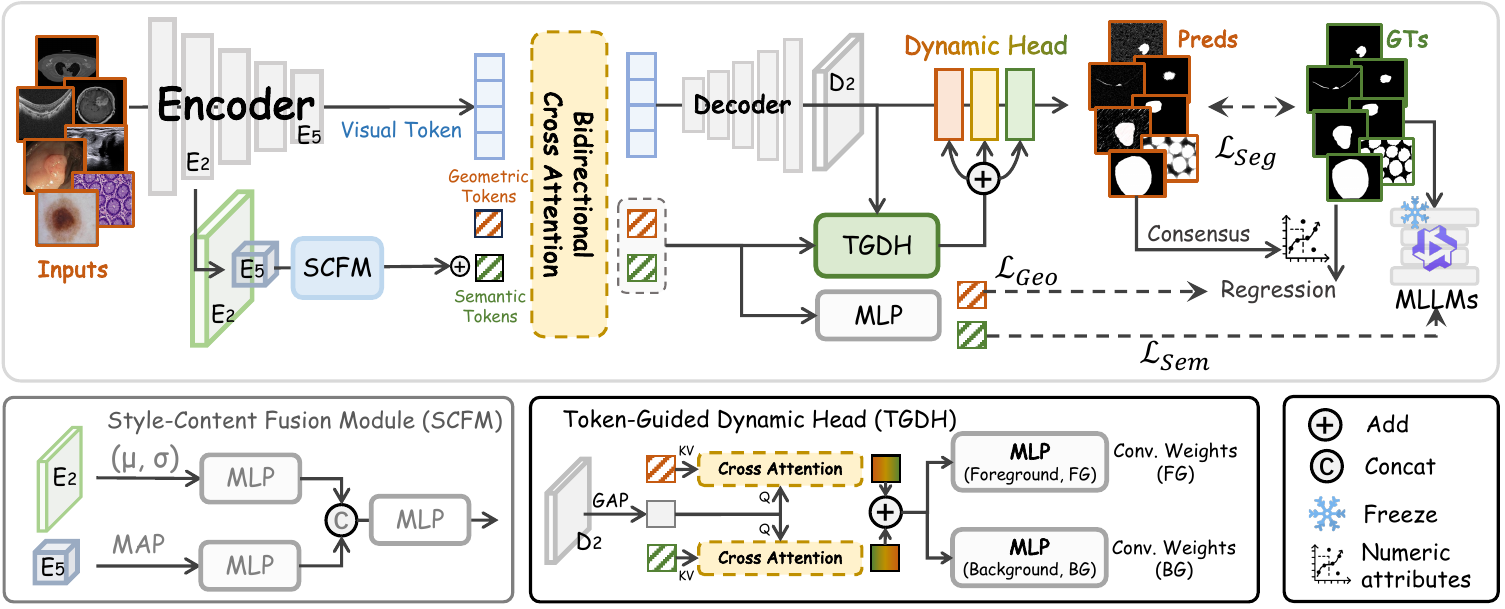} 
  \caption{\textbf{Overview of the proposed Concept-to-Pixel (C2P) framework.} 
  (1) The model first extracts shallow (E2) and deep (E5) backbone features to infer modality embeddings via the Style-Content Fusion Module (SCFM), which are dynamically injected into the \texttt{[SEM]}. (2) The universal \texttt{[GEO]} and modality-aware \texttt{[SEM]} interact with visual tokens through bidirectional Cross Attention. (3) The aggregated concepts are then fed into the Token-Guided Dynamic Head (TGDH) to generate instance-specific parameters for the final Dynamic Head to predict precise masks. To guarantee explicit decoupling, the tokens are deeply supervised by physical properties ($\mathcal{L}_{Geo}$) and MLLM-distilled knowledge ($\mathcal{L}_{Sem}$).}
  \label{fig:framework}
\end{figure}

We propose Concept-to-Pixel (C2P), a universal segmentation framework that explicitly disentangles anatomical understanding into Geometric and Semantic representations to mitigate negative transfer across modalities. Unlike conventional U-Nets that implicitly entangle these features, C2P leverages a deeply guided token-interaction mechanism. As illustrated in \cref{fig:framework}, our framework consists of four key components: (1) {Token Formulation \& Modality Injection}: Initializes universal \texttt{[GEO]} and MLLM-distilled \texttt{[SEM]}, utilizing a Style-Content Fusion Module (SCFM) to resolve cross-modality ambiguities. (2) {Bidirectional Token-Image Interaction}: Enables deep interaction via Cross Attention, where tokens query image features while simultaneously guiding them with high-level priors. (3) {Token-Guided Dynamic Head (TGDH)}: Synthesizes instance-specific convolutional kernels driven by the aggregated tokens and global context for precise mask prediction. (4) {Geometry-Aware Inference Consensus}: An intrinsic evaluation mechanism leveraging the physical self-consistency of \texttt{[GEO]} to filter unreliable predictions.

\subsection{Disentangled Concepts to Pixel-Level Segmentation}
\label{sec:c2p_pipeline}
The core philosophy of our framework is to translate abstract, modality-agnostic physical and clinical expectations into specific pixel-level decision boundaries. To achieve this, we design a three-stage process: Disentangled Concept Extraction, Bidirectional Token-Image Interaction, and Token-Guided Dynamic Prediction.

\vspace{0.5em}
\noindent\textbf{Disentangled Token Formulation.}
To bridge the gap between low-level visual features and high-level medical concepts, we define two distinct sets of tokens: {(i) Geometry Tokens \texttt{[GEO]} ($\mathcal{T}_{geo}$).}
To capture the universal physical attributes of lesions, we initialize a set of $N_{geo}$ learnable tokens $\mathcal{T}_{geo} \in \mathbb{R}^{N_{geo} \times C}$. Crucially, instead of treating these tokens as latent black boxes, each token is explicitly and strictly supervised to regress one of the $N_{geo}$ specific geometric properties. This explicit structural constraint forces the model to encode modality-agnostic ``shape'' and ``location'' physics, ensuring that a topologically identical lesion is recognized consistently regardless of its imaging modality. (ii) Semantic Tokens \texttt{[SEM]} ($\mathcal{T}_{sem}$).
To inject medical semantic knowledge, we leverage the reasoning capability of Large Multi-modal-Language Models, generating structured clinical descriptions for training images across $N_{sem}$ explicit diagnostic dimensions. These texts are encoded into embeddings via PubMedBERT~\cite{gu2021domain}. We initialize $N_{sem}$ learnable \texttt{[SEM]} $\mathcal{T}_{sem} \in \mathbb{R}^{N_{sem} \times C}$ and align them with the MLLM-derived text embeddings using a contrastive loss. This explicit cross-modal alignment effectively distills the structured clinical semantics into the network's latent space, enabling the model to segment targets based on expert-level diagnostic concepts rather than mere pixel intensities.

\vspace{0.5em}
\noindent\textbf{Style-Content Fusion Module (SCFM).}
While geometry is essentially universal in segmentation, semantics are highly modality-dependent. To disentangle semantics, we introduce the Style-Content Fusion Module (SCFM). Specifically, we extract the channel-wise mean $\mu$ (\ie, representing brightness/tone) and standard deviation $\sigma$ (\ie, representing texture/noise) from the shallow backbone features ($E2$) as the Style statistics. Concurrently, we extract the global context from the deep features ($E5$) via global average pooling (GAP) as the Content. These representations are compressed and fused via a gated MLP to infer a dense modality embedding $V_{modality}$. We then selectively inject this modality bias \emph{only} into the \texttt{[SEM]}:
\begin{equation}
    \widehat{\mathcal{T}}_{sem} = \mathcal{T}_{sem} + V_{modality}.
\end{equation}

\vspace{0.5em}
\noindent\textbf{Bidirectional Token-Image Interaction.}
To establish a deep mapping between the abstract concepts and the concrete visual evidence, we propose a multi-layer bidirectional interaction mechanism. First, the deepest visual features from the backbone ($F_{E5}$) are flattened and projected to the token dimension $d$. To preserve spatial topology, a learnable 2D positional encoding is added to $F_{E5}$. Concurrently, the geometric and modality-injected \texttt{[SEM]} are concatenated as $\mathcal{T} = [\mathcal{T}_{geo}, \widehat{\mathcal{T}}_{sem}] \in \mathbb{R}^{(N_{geo}+N_{sem}) \times d}$. These representations then undergo an iterative, two-phase Cross Attention process:

\vspace{0.5em}
\noindent\textbf{Token-to-Image Attention (Abstraction).}
In this phase, the concept tokens act as queries ($Q$) to scan the global image features ($K, V$). This allows each token to adaptively aggregate relevant instance-specific visual evidence from the image (\eg, the ``margin'' token searches for edge-like features). Structurally, this is implemented via Multi-head Cross-Attention (MCA) and a Feed-Forward Network (FFN) with residual connections:
\begin{align}
    \mathcal{T}' &= \mathcal{T} + \text{MCA}(\text{LN}(\mathcal{T}), F_{E5}, F_{E5}), \\
    \mathcal{T}_{out} &= \mathcal{T}' + \text{FFN}(\text{LN}(\mathcal{T}')),
\end{align}
where $\text{LN}(\cdot)$ denotes Layer Normalization operations. Through this process, the concept tokens transition from static global representations to dynamic, instance-aware concept representations.

\vspace{0.5em}
\noindent\textbf{Image-to-Token Attention (Guidance).}
Conversely, the image features must be refined by the high-level concepts to suppress irrelevant background noise. Here, the image features act as queries to retrieve information from the updated tokens dictionary:
\begin{align}
    F'_{E5} &= F_{E5} + \text{MCA}(\text{LN}(F_{E5}), \mathcal{T}_{out}, \mathcal{T}_{out}), \\
    F_{out} &= F'_{E5} + \text{FFN}(\text{LN}(F'_{E5})).
\end{align}
By querying the concept tokens, the pixel-level features explicitly highlight regions that align with the predefined geometric and semantic expectations, yielding deeply coupled representations.

\vspace{0.5em}
\noindent\textbf{Token-Guided Dynamic Head (TGDH).}
To overcome the rigidity of static segmentation heads, which apply fixed weights across highly heterogeneous modalities, we propose the Token-Guided Dynamic Head (TGDH). This mechanism synthesizes image-specific convolutional parameters directly from the disentangled concept tokens. First, the refined image features undergo a decoder pathway to produce high-resolution representations $D_2$. We apply global average pooling to $D_2$ to obtain a global image query $Q_{img}$. This query is then fed into two parallel Cross-Attention modules to aggregate the most relevant insights from the final \texttt{[GEO]} and \texttt{[SEM]}:
\begin{equation}
    \mathcal{F}_{geo} = \text{CrossAttn}(Q_{img}, \mathcal{T}_{geo, out}), \quad \mathcal{F}_{sem} = \text{CrossAttn}(Q_{img}, \mathcal{T}_{sem, out}).
\end{equation}
The aggregated token features and the global image query are concatenated and processed by a Kernel Generator MLP ($\Psi$) to synthesize dynamic parameters for both foreground ($fg$) and background ($bg$) convolutions:
\begin{equation}
  \mathcal{W}_{fg}, \mathcal{W}_{bg} = \Psi(\text{Concat}(\mathcal{F}_{geo}, \mathcal{F}_{sem}, Q_{img})).
\end{equation}
Finally, the segmentation masks are generated by the Dynamic Head, which applies these dynamically generated convolutional weights $\mathcal{W}$ directly onto the high-resolution feature map $D_2$. This ensures that the pixel-level decision boundary is dynamically governed by explicit physical and clinical reasoning.

\subsection{Multi-Task Supervision Strategy}
To ensure the tokens genuinely learn the disentangled representations, we optimize the network using a multi-task objective $\mathcal{L}_{total} = \lambda_{seg}\mathcal{L}_{seg} + \lambda_{geo}\mathcal{L}_{geo} + \lambda_{sem}\mathcal{L}_{sem}$.

\begin{itemize}
    \item \textbf{Segmentation Loss ($\mathcal{L}_{seg}$):} We employ a combination of a Boundary-Aware Structure Loss ($\mathcal{L}_{struct}$) and a Dice Loss ($\mathcal{L}_{dice}$) for both foreground ($fg$) and background ($bg$) predictions:
    \begin{equation}
        \mathcal{L}_{seg} = \sum_{c \in \{fg, bg\}} \big( \mathcal{L}_{struct}(P_c, M_c) + \mathcal{L}_{dice}(P_c, M_c) \big),
    \end{equation}
    where $P_c$ and $M_c$ denote the predicted probabilities and ground-truth masks for class $c$. To compute $\mathcal{L}_{struct}$, we extract lesion boundaries by computing the absolute difference between the mask and its average-pooled version, assigning dynamic spatial weights $W$ to hard-to-segment edge regions:
    \begin{equation}
        W = 1 + 5 \cdot |\text{AvgPool}_{31\times31}(M_c) - M_c|.
    \end{equation}

    This weight map is synergistically applied to emphasize boundary regions. To further address class imbalance and focus on hard-to-classify pixels, we combine this spatial weight with a Focal Loss ($\gamma=2.0$) and an Intersection over Union (IoU) loss, formulating the structure loss as:
    \begin{equation}
        \mathcal{L}_{struct} = \mathcal{L}_{Focal}^{W}(P_c, M_c) + \mathcal{L}_{IoU}^{W}(P_c, M_c).
    \end{equation}

    \item \textbf{Geometric Regression Loss ($\mathcal{L}_{geo}$):} We apply the Mean Squared Error (MSE) loss across all $N_{geo}$ geometric properties (including bounding box, area, perimeter, \etc). This strictly supervises the \texttt{[GEO]} to capture the physical ground truth:
    \begin{equation}
        \mathcal{L}_{geo} = \frac{1}{N_{geo}} \sum_{i=1}^{N_{geo}} \text{MSE}(\mathcal{T}_{geo}^{(i)}, g^{(i)}),
    \end{equation}
    where $g^{(i)}$ represents the normalized ground-truth scalar for the $i$-th geometric property.
    
    \item \textbf{Semantic Alignment Loss ($\mathcal{L}_{sem}$):} To inject abstract medical knowledge, we compute the Cosine Embedding Loss between the learned \texttt{[SEM]} and the pre-computed text embeddings derived from large multi-modal models:
    \begin{equation}
        \mathcal{L}_{sem} = \frac{1}{N_{sem}} \sum_{j=1}^{N_{sem}} \big( 1 - \cos(\mathcal{T}_{sem}^{(j)}, E_{text}^{(j)}) \big),
    \end{equation}
    where $E_{text}^{(j)}$ is the structured clinical embedding for the $j$-th diagnostic dimension. This forces the latent space to seamlessly align with expert medical semantics.
\end{itemize}

\subsection{Geometry-Aware Inference Consensus}
 Standard inference paradigms often treat all predictions equally, making them vulnerable to ambiguous boundaries or severe imaging artifacts. To enhance the robustness of our framework, we propose a Geometry-Aware Inference Consensus mechanism that utilizes the explicitly learned \texttt{[GEO]} as intrinsic quality inspectors.
\begin{wrapfigure}{r}{0.57\textwidth}
  \begin{center}
  \vspace{-8mm}
    \includegraphics[width=0.55\textwidth]{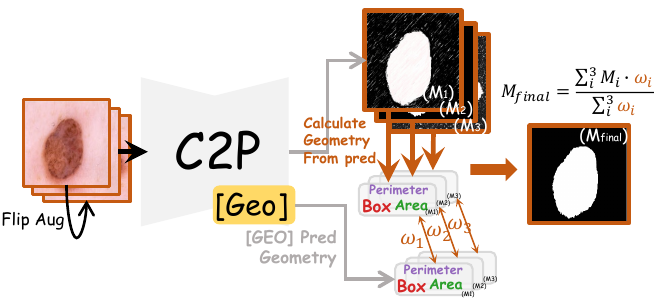}
  \end{center}
  \vspace{-4mm}
  \caption{\textbf{Workflow of Geometry-Aware Inference Consensus.} \texttt{[GEO]} compares regressed geometry with mask-derived geometry to weight each view and aggregate a robust final prediction.\label{fig:consensus}}
  \vspace{-6mm}
\end{wrapfigure}
As shown in \cref{fig:consensus}, our core insight relies on geometric self-consistency: a reliable prediction must satisfy strict physical constraints, where both the pixel-level area and spatial location of the generated mask tightly match the abstract, regression-level expectations predicted by the \texttt{[GEO]}. During inference, given multiple perturbed views $x_i$ of an input image, the model outputs the segmentation mask $M_i$, along with a regressed area $a_i^{reg}$ and a centroid $c_i^{reg}$. We then calculate the actual pixel-wise area $a_i^{px}$ and centroid $c_i^{px}$ directly from $M_i$. An inconsistency penalty is computed based on both scale and location shifts, and the confidence weight $w_i$ for each view is derived as:
\begin{equation}
    w_i = \exp\left(- \lambda \cdot (10 \cdot |a_i^{reg} - a_i^{px}| + \|c_i^{reg} - c_i^{px}\|_2)\right),
\end{equation}
where $\lambda$ controls the penalty sharpness. This weighting strategy heavily penalizes predictions where the network's abstract geometric reasoning contradicts its concrete visual output. Furthermore, an explicit false-positive suppression is applied to masks with contradicting near-zero predicted areas. The final prediction is dynamically aggregated as $M_{final} = (\sum w_i M_i) / (\sum w_i)$, effectively filtering out unreliable outliers based purely on multidimensional physical self-consistency.

\section{Experiments}
\label{sec:experiments}

\subsection{Datasets and Implementation Details}
\label{sec:exp_setup}

\noindent\textbf{Datasets.} Aiming to conduct fair comparisons with the previous works~\cite{chang2025unified}, we conduct our experiments on the same 8 public datasets, spanning 7 distinct imaging modalities. This comprehensive benchmark includes AMD-SD~\cite{hu2024amd} (OCT), BTD~\cite{cheng2015enhanced,cheng2016retrieval} (Brain MRI), EBHI-Seg~\cite{shi2023ebhi} (Pathology), TNUI-2021~\cite{zhou2022h} and BUSI~\cite{al2020dataset} (Ultrasound), Colon Polyp~\cite{bernal2015wm,jha2019kvasir,silva2014toward,tajbakhsh2015automated,vazquez2017benchmark} (Endoscopy), COVID-19~\cite{fan2020inf} (CT), and ISIC 2018~\cite{codella2019skin} (Dermoscopy). Furthermore, to comprehensively evaluate the zero-shot generalization capability of our model, we extend our evaluation to 8 additional datasets. Specifically, for unseen datasets within seen modalities, we utilize TN3K~\cite{gong2021multi} and BUSBRA~\cite{gomez2024bus} (Ultrasound), PH2~\cite{mendoncca2015ph2} (Dermoscopy), ACDC~\cite{bernard2018deep} (MRI), and GlaS~\cite{sirinukunwattana2017gland} (Pathology). For entirely unseen modalities, we test on Montgomery~\cite{jaeger2014two} and Covidquex~\cite{tahir2021covid} (X-Ray), as well as ISBI EM~\cite{cardona2010integrated} (Electron Microscopy). Details of datasets are provided in the Appendix.

\begin{table*}[tb]
  \caption{Comparison of different methods on multiple datasets (Dice \%). The abbreviations Path., Endo., and Dermo. correspond to Pathology, Endoscopy, and Dermoscopy, respectively. The best results are highlighted in \textbf{bold}, and the second-best results are \underline{underlined}.}
  \label{tab:comparison}
  \centering
  \setlength{\tabcolsep}{3pt}
  \resizebox{\textwidth}{!}{%
  \begin{tabular}{@{}lccccccccc@{}}
    \toprule
    \multirow{2}{*}{Methods} & Path. & \multicolumn{2}{c}{Ultrasound} & Endo. & CT & Dermo. & OCT & MRI  & \multirow{2}{*}{Avg} \\
    \cmidrule{3-4}
     & \makebox[1.4cm][c]{EBHI} & \makebox[1.4cm][c]{TNUI} & \makebox[1.4cm][c]{Breast} & \makebox[1.4cm][c]{Polyp} & \makebox[1.4cm][c]{COVID} & \makebox[1.4cm][c]{ISIC2018} & \makebox[1.4cm][c]{AMDSD} & \makebox[1.4cm][c]{BTD} &  \\
    \midrule
    
    \multicolumn{10}{c}{\textcolor{gray}{\textit{Specialized Models (One model for one task)}}} \\
    \midrule
    UNet (MICCAI'25)~\cite{ronneberger2015u} & 94.61 & 85.11 & 73.00 & 87.66 & 78.62 & 87.76 & 85.57 & 80.92 & 84.16 \\
    AURA-Net (ISBI'21)~\cite{cohen2021aura} & 95.08 & 87.55 & 79.34 & 90.53 & 79.28 & 89.92 & 85.91 & 84.75 & 86.54 \\
    UTANet (AAAI'25)~\cite{nguyen2024ac} & 94.82 & 86.79 & 78.78 & 89.75 & 79.14 & 89.32 & 85.72 & 83.57 & 85.99 \\
    Swin-UMamba (MICCAI'24)~\cite{liu2024swin} & 95.14 & 87.86 & 81.35 & 91.42 & 82.07 & 90.40 & 85.51 & 84.80 & 87.32 \\
    LGMSNet (ECAI'25)~\cite{dong2025lgmsnet} & 95.07 & 86.40 & 79.80 & 89.06 & 79.92 & 90.02 & 85.02 & 83.63 & 86.12 \\
    MEGANet (CVPR'25)~\cite{bui2024meganet} & 95.06 & 87.96 & 80.30 & 91.29 & 81.70 & 90.45 & 85.89 & 84.73 & 87.17 \\
    RWKV-UNet (arXiv'25)~\cite{ye2025u} & 95.26 & 88.21 & \underline{81.40} & 91.41 & 81.41 & 90.55 & 85.65 & 85.15 & \underline{87.38} \\
    nnUNetV2 (NM'21)~\cite{isensee2021nnu} & 95.26 & \underline{88.22} & 78.20 & 90.47 & 79.44 & 87.54 & \textbf{88.01} & \textbf{86.32} & 86.68 \\
    \midrule
    \multicolumn{10}{c}{\textcolor{gray}{\textit{Universal Models (One model for all tasks)}}} \\
    \midrule
    UniverSeg (CVPR'23)~\cite{butoi2023universeg} & 91.28 & 71.75 & 66.04 & 63.18 & 38.69 & 82.57 & 68.30 & 63.37 & 68.15 \\
    Tyche (CVPR'24)~\cite{rakic2024tyche} & 94.20 & 85.00 & 73.39 & 83.47 & 74.18 & 88.24 & 82.44 & 79.34 & 82.53 \\
    Spider (ICML'24)~\cite{zhao2024spider} & \underline{95.29} & 88.08 & 81.11 & \underline{92.40} & \underline{83.14} & \underline{90.67} & 83.21 & 84.66 & 87.32 \\
    SR-ICL (CVPR'25)~\cite{chang2025unified} & 94.78 & 87.75 & 80.10 & 92.22 & 82.92 & 89.99 & 84.88 & 85.37 & 87.25 \\
    \rowcolor{gray!15} \textbf{C2P (ours)} & \textbf{95.44} & \textbf{88.80} & \textbf{81.70} & \textbf{93.13} & \textbf{83.91} & \textbf{90.79} & \underline{86.27} & \underline{85.73} & \textbf{88.22} \\
    \bottomrule
  \end{tabular}%
  }
\end{table*}

\begin{figure}[tb]
  \centering
  \includegraphics[width=0.95\textwidth]{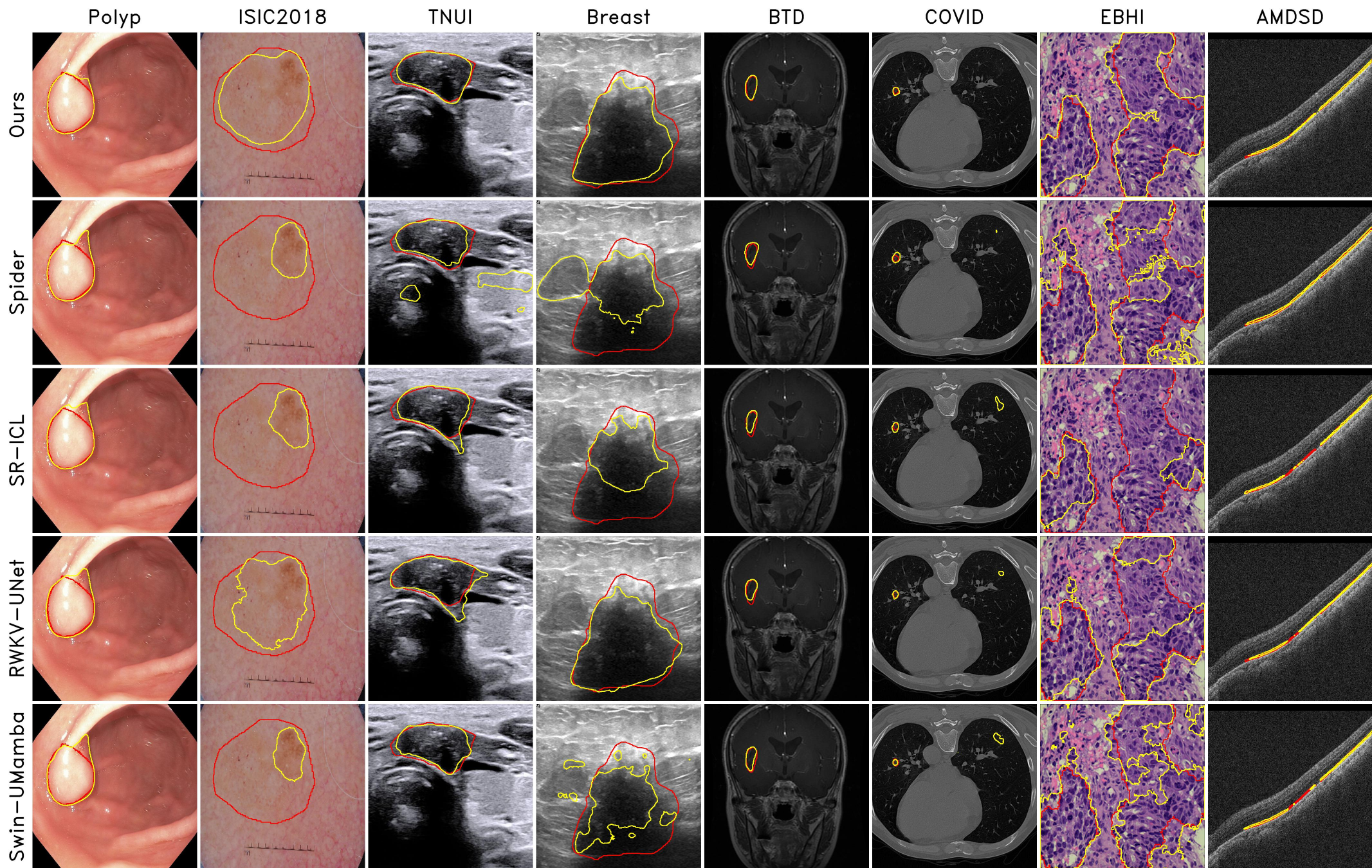} 
  \caption{\textbf{Qualitative comparison on representative challenging cases.} From left to right: Colon Polyp, ISIC2018, TNUI, Breast, BTD, COVID, EBHI, and AMDSD. The \textcolor{red}{red contours} indicate the Ground Truth masks, and the \textcolor{yellow}{yellow contours} denote the model predictions.}
  \label{fig:qualitative_results}
\end{figure}

\noindent\textbf{Implementation Details.} 
To ensure fair and rigorous comparisons, we strictly conduct same experimental settings (\eg, data splits, data augmentations, and evaluation criteria).
We implemented our framework in PyTorch~\cite{paszke2019pytorch} with ConvNeXt-B~\cite{liu2022convnet,woo2023convnext} as the backbone, setting the input resolution to $384 \times 384$. The model is trained for 50 epochs using the Adam~\cite{kingma2014adam} optimizer with a batch size of 8 and a cosine decayed initial learning rate of $1e-4$. To provide explicit geometric supervision, 9 physical attributes for \texttt{[GEO]} (\ie, bounding box, area, perimeter, aspect ratio, compactness, centroid, eccentricity, orientation, and solidity) are pre-computed and extracted offline directly from the ground-truth (GT) masks. For semantic knowledge, we employ Qwen3-VL-Plus~\cite{bai2023qwen,wang2024qwen2,bai2025qwen3} to generate structured medical reports across 9 explicit clinical dimensions (\ie, morphology, margin definition, internal texture, surrounding interaction, boundary distinctness, malignancy risk, pathological inference, differential reasoning, and predicted diagnosis). These reports are subsequently encoded into 768-dimensional token embeddings using PubMedBERT~\cite{gu2021domain}. Detailed offline knowledge pre-processing and configurations are provided in the Appendix.

\noindent\textbf{Comparison Baselines.} We compare our method with comprehensive baselines divided into two categories: (1) Specialized segmentation models:  UNet~\cite{ronneberger2015u}, nnU-NetV2~\cite{isensee2021nnu}, and a curated selection of the most performant models according to U-Bench~\cite{tang2025u}: RWKV-UNet~\cite{ye2025u}, AURA-Net~\cite{cohen2021aura}, UTANet~\cite{nguyen2024ac}, MEGANet~\cite{bui2024meganet}, Swin-UMamba~\cite{liu2024swin}, and LGMSNet~\cite{dong2025lgmsnet}. (2) Universal segmentation models: UniverSeg~\cite{butoi2023universeg}, Tyche~\cite{rakic2024tyche}, Spider~\cite{zhao2024spider}, and SR-ICL~\cite{chang2025unified}.\\

\subsection{Comparison with State-of-the-Arts}
\label{sec:comparison}

\noindent\textbf{Outperforming Specialized and Universal Baselines.}
As shown in \cref{tab:comparison}, our method achieves an average Dice of 88.22\%, surpassing strong specialized method nnU-NetV2 and RWKV-UNet. Compared to universal models, our C2P demonstrates a clear performance leap, outperforming the previous SOTA (Spider) by 0.90\% in average Dice. Unlike existing methods (\eg, UniverSeg and Tyche) that struggle with extreme domain shifts, our approach excels in challenging tasks such as Polyp (+0.73\%) and COVID-19 (+0.77\%) segmentation.
This suggests that our design successfully improves the recognition of diverse textural and morphological variations compared to previous methods. 
Notably, in the modalities with fuzzy and blurred boundaries (e.g. Ultrasound, CT), as shown in \cref{fig:qualitative_results}, our method still predicts smooth, highly accurate, and geometrically plausible boundaries, setting new SOTAs across different datasets. 

\begin{table}[t!]
\centering
\caption{
\textbf{Zero-shot generalization results on unseen datasets and unseen modalities.} C2P is evaluated without any support images, new tokens, or MLLM prompts for unseen data, and compared against specialized (nnUNetv2) and universal baselines (Spider with 64-shot references, SR-ICL with iterative refinement). The best results are highlighted in \textbf{bold}, and the second-best results are \underline{underlined}. C2P achieves competitive performance on unseen datasets and outperforms all baselines on unseen X-ray and Ultrasound modalities, showing its generalization ability. 
}
\label{tab:zero_shot}
\resizebox{\textwidth}{!}{%
\begin{tabular}{lccccc}
\toprule
\multirow{2}{*}{Dataset} & \multirow{2}{*}{Modality}  & nnUNetV2~\cite{isensee2021nnu} & Spider~\cite{zhao2024spider} & SR-ICL~\cite{chang2025unified} & \cellcolor{gray!15}{\textbf{C2P (ours)}} \\ 
\cmidrule{3-6}
& & (Zero-Ref) &(64-Ref) &(Self-Ref)&\cellcolor{gray!15}{(Zero-Ref)} \\

\midrule
\multicolumn{6}{c}{\textcolor{gray}{\textit{Generalization to Unseen Datasets (Seen Modalities)}}} \\ \midrule
TN3K~\cite{gong2021multi} & Ultrasound & 52.61 & \underline{74.24} & 70.59 & \cellcolor{gray!15}{\textbf{75.14}} \\
PH2~\cite{mendoncca2015ph2} & Dermoscopy & 56.39 & 90.47 & \underline{90.70} & \cellcolor{gray!15}{\textbf{90.89}} \\
ACDC~\cite{bernard2018deep} & MRI & 29.17 & \underline{63.26} & 59.95 & \cellcolor{gray!15}{\textbf{71.87}} \\
BUSBRA~\cite{gomez2024bus} & Ultrasound & 78.83 & \textbf{85.16} & 80.70 & \cellcolor{gray!15}{\underline{83.39}} \\
GlaS~\cite{sirinukunwattana2017gland} & Pathology & \textbf{70.92} & \underline{56.42} & 53.78 & \cellcolor{gray!15}{53.99} \\ \midrule

\multicolumn{6}{c}{\textcolor{gray}{\textit{Generalization to Unseen Modalities}}} \\ \midrule
Montgomery~\cite{jaeger2014two} & X-Ray & - & \underline{83.89} & 67.00 & \cellcolor{gray!15}{\textbf{87.60}} \\
Covidquex~\cite{tahir2021covid} & X-Ray & - & \underline{50.52} & 46.80 & \cellcolor{gray!15}{\textbf{53.31}} \\
ISBI EM~\cite{cardona2010integrated} & Electron Microscopy & - & \underline{64.63} & \textbf{83.21} & \cellcolor{gray!15}{26.79} \\ \bottomrule
\end{tabular}%
}
\vspace{-4mm}
\end{table}

\noindent\textbf{Strong Zero-Shot Generalization. \label{sec:generalization}} We evaluate our method under zero-shot inference setting, where no new \texttt{[GEO]}, \texttt{[SEM]}, or external MLLM text prompts are introduced or generated for unseen data. The evaluation is categorized into two scenarios: generalization to unseen datasets (intra-domain shift) and generalization to entirely unseen modalities. We compare our approach against the strongest specialized baseline nnUNetv2 and two competitive universal baselines: Spider, which utilizes an extensive support set of 64 reference images, and SR-ICL with its self-refinement mechanism. 

We first evaluate the models on five unseen datasets (TN3K, PH2, BUSBRA, ACDC, GlaS).  As shown in \cref{tab:zero_shot}, unlike Spider with 64-reference images or SR-ICL with two-stage iterative refinement, our model achieves competitive performance without requiring support images in most modalities (e.g. Dice of 75.15\% in TN3K, 71.87\% in ACDC). This striking success demonstrates that our \texttt{[SEM]} robustly adapt to the varying pixel intensity distributions, and \texttt{[GEO]} successfully capture essential structural priors. 

Then, we evaluate on entirely unseen modalities (Montgomery X-ray~\cite{jaeger2014two}, Covidquex and ISBI Electron Microscopy (ISBI EM)~\cite{cardona2010integrated}). On the unseen X-ray modality dataset Montgomery and Covidquex, SR-ICL suffers a catastrophic failure, dropping to 67.00\% and 46.80\% Dice. Due to its self-refinement, when faced with a completely unseen modality, the massive domain gap renders the reference images ineffective. In contrast, our Concept-to-Pixel achieves a remarkable result, with Dice score of 87.6\% and 53.31\%. We argue that \texttt{[GEO]} show effectiveness in generalization: by learning compact shape priors across diverse lesion types during training, the model can recognize and segment unseen lesions that share similar geometries, even across previously unseen modalities.
Conversely, our model struggles on the ISBI EM dataset (26.79\% Dice). Since all 8 training datasets consist predominantly of macroscopic, compact structures (\eg, polyps, nodules, tumors), the \texttt{[GEO]} have been trained to encode a strong compactness prior. This prior is incompatible with the filamentous, highly elongated topology of neuronal structures in EM images, causing geometric supervision to actively conflict with the correct prediction. This failure case is consistent with the design intent of C2P: the explicit geometric constraints are effective precisely because they are strict, and will generalize poorly when the target topology fundamentally violates the training distribution.

\subsection{Ablation Studies}
\label{sec:ablation}

To investigate the contribution of each component in our \textbf{Concept-to-Pixel} framework, we conduct extensive ablation studies on the combined 8-dataset benchmark and report the average Dice scores. 
Our ablations start with a standard UNet with
ConvNeXtV2-based backbone (ID 1).

\begin{table}[t!]
\centering
\caption{\textbf{Ablation studies of Concept-to-Pixel.} (Static): Linear Segmentation Head. The best results are highlighted in \textbf{bold}.  
}
\label{tab:ablation_component}
\begin{tabular}{cccccccc}
\toprule
\multirow{2}{*}[-0.8ex]{ID} & 
\multirow{2}{*}[-0.8ex]{\begin{tabular}[c]{@{}c@{}}Geometry \\ Tokens\end{tabular}} & 
\multirow{2}{*}[-0.8ex]{\begin{tabular}[c]{@{}c@{}}Semantic \\ Tokens\end{tabular}} & 
\multirow{2}{*}[-0.8ex]{\begin{tabular}[c]{@{}c@{}}Dynamic \\ Head\end{tabular}} & 
\multirow{2}{*}[-0.8ex]{\begin{tabular}[c]{@{}c@{}}Inference \\ Strategy\end{tabular}} & 
\multicolumn{2}{c}{Partial Datasets (\%)} & 
\multirow{2}{*}[-0.8ex]{\begin{tabular}[c]{@{}c@{}}Avg Dice \\ (\%)\end{tabular}} \\
\cmidrule(lr){6-7}
 & & & & & AMDSD & COVID &  \\ \midrule
1 & - & - & - (Static) & None & 82.19 & 76.27 & 82.38 \\
2 & - & \checkmark & - (Static) & None & 84.93 & 80.27 & 87.15 \\
3 & \checkmark & - & - (Static) & None & 84.98 & 82.66 & 87.31 \\
4 & - & - & \checkmark & None & 84.89 & 82.45 & 87.19 \\
5 & \checkmark & - & \checkmark & None & 85.35 & 81.34 & 87.29 \\
6 & - & \checkmark & \checkmark & None & 85.00 & 82.27 & 87.43 \\ \midrule
7 & \checkmark & \checkmark & \checkmark & None & 86.07 & 83.49 & 87.79 \\
8 & \checkmark & \checkmark & \checkmark & Standard TTA & 86.20 & 83.60 & 88.16 \\
9 & \checkmark & \checkmark & \checkmark & {Geometry-Aware} & \textbf{86.27} & \textbf{83.91} & \textbf{88.22} \\ \bottomrule
\end{tabular}%
\end{table}

\noindent\textbf{Impact of Disentangled Concepts.}
The most significant performance leap comes from the introduction of explicit concept tokens. 
As shown in \cref{tab:ablation_component}, the baseline model (ID 1) struggles with an Dice of 82.38\%, suffering from negative transfer across diverse modalities. 
However, simply injecting \texttt{[SEM]} (ID 2) or \texttt{[GEO]} (ID 3) boosts the Dice to 87.15\% and 87.31\%, respectively. 
Remarkably, the static head guided by \texttt{[GEO]} (ID 3, 87.31\%) even outperforms the blind Dynamic Head (ID 3 and 4, 87.19\%). 
This strongly validates our core hypothesis: \emph{explicit medical knowledge (what to look for and where) is more critical than complex architectural changes alone.}

\noindent\textbf{Synergy of Dynamic Convolution and Dual Tokens.}
While tokens provide strong priors, the static segmentation head limits adaptability. 
By employing Dynamic Convolution conditioned on both token types (ID 7), we achieve a further improvement to 87.79\%. 
Comparing ID 7 against single-token variants (ID 5 and 6), the combination of geometric constraints and semantic reasoning yields a synergistic effect (+0.50\% Dice over ID 5), proving that "shape" and "texture" are complementary information streams that the dynamic head can effectively utilize to synthesize instance-specific weights.

\noindent\textbf{Effectiveness of Geometry-Aware Consistency.}
Finally, we analyze the inference strategy. 
Standard Test-Time Augmentation~\cite{shanmugam2021better} (ID 8), which averages predictions from augmented views, improves Dice to 88.16\%. 
Our Geometry-Aware Consistency (ID 9) further pushes the performance to 88.22\%. 
Unlike blind averaging, our method utilizes the regressed geometric attributes to penalize inconsistent predictions (e.g., when the predicted mask area contradicts the token's expectation). 
This demonstrates that the disentangled geometric knowledge serves not only as a training constraint but also as a reliable "self-check" mechanism during deployment.

\subsection{Representation and Generalization Analysis}
\label{sec:tsne_vis}

We further investigate the functional impact of our \texttt{[SEM]} and \texttt{[GEO]} by visualizing their feature spaces and dependency structures using t-SNE and attention maps, respectively. Our findings are summarized in this section.

\begin{figure*}[t]
  \centering
  \includegraphics[width=1\textwidth]{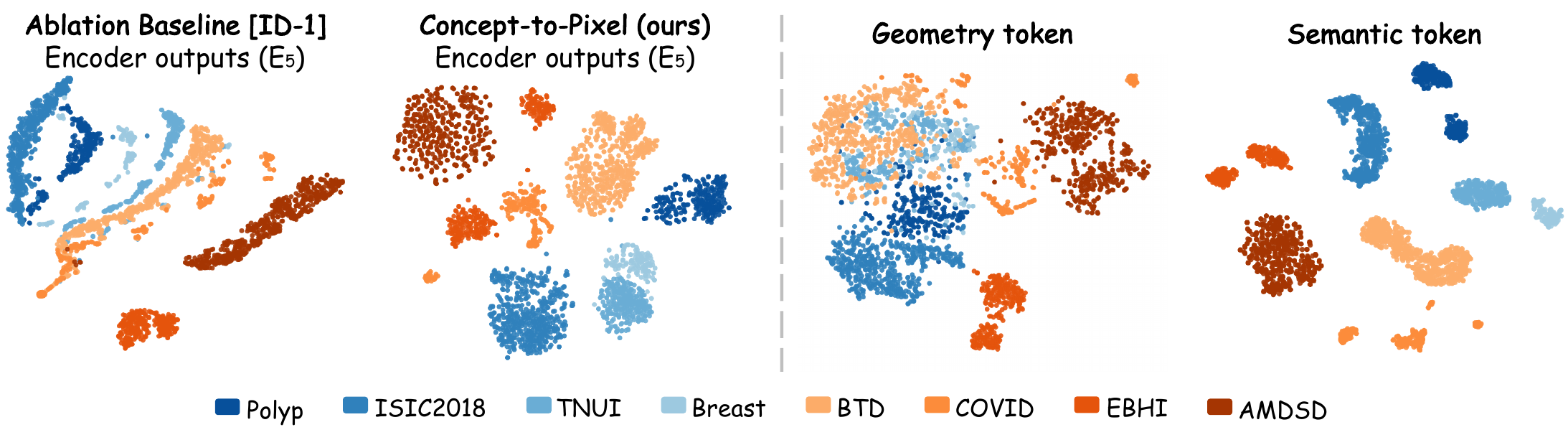} 
  \vspace{-4mm}
  \caption{\textbf{In-domain representation analysis.} \textbf{(Left) t-SNE of encoder outputs (E5 features).} Without C2P, features from different datasets show entanglements. Our C2P effectively separates the features. \textbf{(Right) t-SNE of \texttt{[GEO]} and \texttt{[SEM]}.} Some geometry tokens are clustered, since samples from different dataset may have similar geometry attributes. In contrast, semantic tokens are clearly separated, indicating that each token has successfully encoded discriminative semantic concepts tied to its corresponding domain.}
  \vspace{-4mm}
  \label{fig:tsne_combined}
\end{figure*}

\noindent\textbf{In-Domain Representation Analysis.} As shown in \cref{fig:tsne_combined} (Left), we first visualize the deepest backbone representations (\ie, E5) when the \texttt{[GEO]} and \texttt{[SEM]} supervision is removed (same as the configuration of the ablation ID 1). It shows that, lacking explicit concept guidance, the pure backbone features suffer from severe domain bias. In contrast, when introducing our C2P, the network intrinsically learns more cohesive modality representations. Furthermore, we provide t-SNE visualizations for both \texttt{[SEM]} and \texttt{[GEO]} in \cref{fig:tsne_combined} (Right) to investigate their latent structures. The \texttt{[GEO]} exhibit an entangled and overlapping distribution across different domains, effectively verifying their capacity to extract modality-invariant features and universally shared concepts. In contrast, the \texttt{[SEM]} are distilled into highly compact and distinct clusters with minimal intra-class variance. This clear separation indicates that the \texttt{[SEM]} successfully capture modality-specific characteristics.

\begin{figure}[t]
    \centering
    \includegraphics[width=\linewidth]{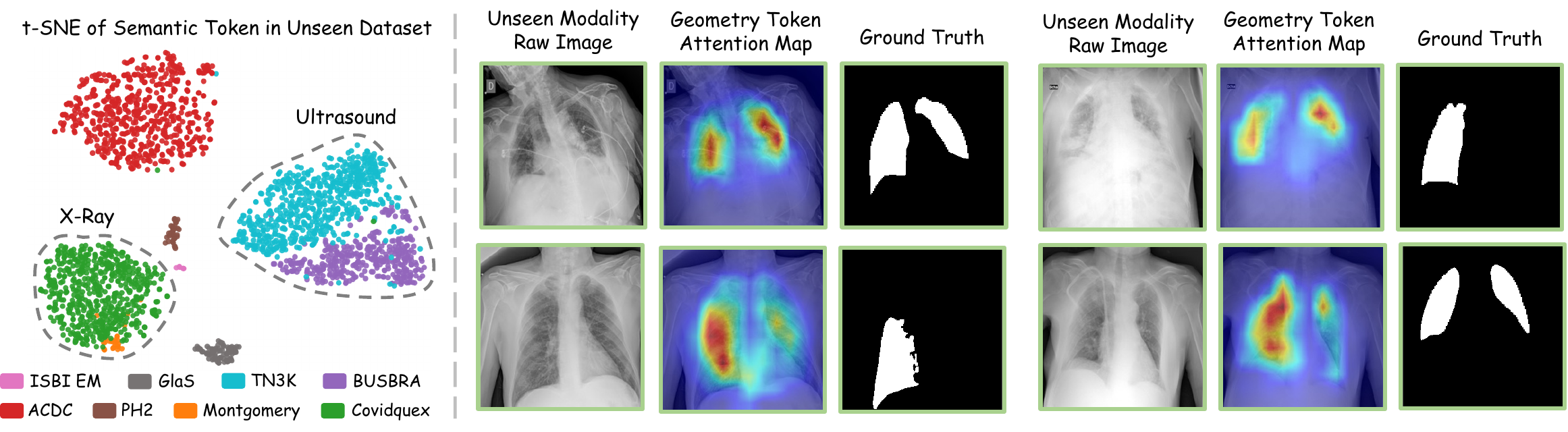}
    \caption{\textbf{Generalization Capability Analysis.} \textbf{(Left) t-SNE analysis of \texttt{[SEM]} in unseen dataset.} Although our model has never seen the datasets, the result shows clear separation among modalities, while different datasets also belong to different clusters. \textbf{(Right) Visualization of the attention map between \texttt{[GEO]} and visual tokens.} Our \texttt{[GEO]} shows an attention pattern that is highly correlated with the ground truth mask.}
    \label{fig:zero-shot}
    \vspace{-4mm}
\end{figure}

\noindent\textbf{Generalization Capability Analysis.}
We evaluate C2P's generalization by visualizing \texttt{[SEM]} and \texttt{[GEO]} on unseen datasets via t-SNE and attention maps, respectively.
As shown in \cref{fig:zero-shot} (Left), \texttt{[SEM]} form cohesive clusters on unseen data. Notably, samples from different sources within the same modality (BUS-BRA and TN3K, both Ultrasound) are mapped to the same latent region. Even for a completely unseen modality (X-Ray), \texttt{[SEM]} maintain clear clustering, suggesting that C2P develops generalizable modality-aware representations beyond its training distribution.
As shown in \cref{fig:zero-shot} (Right), \texttt{[GEO]} accurately localize task-relevant regions on the Covidquex dataset (COVID-19 X-Ray, a completely unseen modality). The attention maps show that \texttt{[GEO]} simultaneously captures global lung structure and concentrates on infected regions, demonstrating that the learned geometric priors transfer effectively to unseen modalities.

\section{Conclusion}
\label{sec:conclusion}

In this paper, we presented \textbf{Concept-to-Pixel (C2P)}, a novel prompt-free universal medical image segmentation framework. To overcome the long-standing challenge of negative transfer across heterogeneous imaging modalities, C2P explicitly disentangles anatomical reasoning into modality-agnostic \texttt{[GEO]} and MLLMs-distilled \texttt{[SEM]}. By inferring and injecting modality awareness solely into the semantic space and leveraging a bidirectional token-image transformer, our model generates instance-specific convolutional kernels driven entirely by explicit physical and clinical concepts. Furthermore, we introduced an intrinsic {Geometry-Aware Inference Consensus} mechanism to confidently filter unreliable predictions. Extensive experiments demonstrate that C2P not only successfully suppresses negative transfer but remarkably outperforms state-of-the-art universal methods  and strong task-specialized baselines. Notably, our framework exhibits exceptional zero-shot generalization to completely unseen macroscopic modalities without requiring external reference images. While the current formulation naturally internalizes a topological inductive bias towards compact lesions, expanding the training mixture to encompass filamentous structures (\eg, vessels, neurons) presents a straightforward and promising avenue for future work. Ultimately, Concept-to-Pixel paves a highly interpretable, robust, and generalizable path toward true foundation models in medical image analysis.

\bibliographystyle{splncs04}
\bibliography{main}

\appendix
\nolinenumbers
\input{Appendix.tex}

\end{document}

%% file: Appendix.tex
\newpage
\title{Concept-to-Pixel: Prompt-Free Universal Medical Image Segmentation} 

\vspace{25mm}
\begin{center}
    \large  \textbf{Supplementary Material}
\end{center}
\vspace{-5mm}
\section{Datasets}  
\label{sec:datasets_appendix}

To comprehensively evaluate the universality, robustness, and zero-shot generalization capability of our proposed Concept-to-Pixel (C2P) framework, we construct a large-scale benchmark comprising 16 public medical image segmentation datasets. These datasets span 9 distinct imaging modalities and cover a wide range of anatomical structures and lesion types. The detailed characteristics of all datasets are summarized in \cref{tab:all_datasets}.

\begin{table}[h]
\centering
\caption{\textbf{Summary of medical datasets evaluating in the experiments.} The benchmark is carefully carefully divided into in-domain training/evaluation datasets and zero-shot generalization datasets. Zero-shot datasets are  categorized by whether their imaging modalities are seen during training.}
\label{tab:all_datasets}
\resizebox{\textwidth}{!}{%
\begin{tabular}{l|l|l|l}
\toprule
\textbf{Experimental Phase} & \textbf{Modality} & \textbf{Dataset} & \textbf{Segmentation Target} \\ \midrule

\multirow{8}{*}{\begin{tabular}[c]{@{}l@{}}\textbf{In-Domain Phase}\\ (Joint Training \&\\ In-Domain Evaluation)\end{tabular}} 
& OCT & AMD-SD~\cite{hu2024amd} & Retinal Lesion (Wet AMD) \\
& MRI (T1) & BTD~\cite{cheng2015enhanced,cheng2016retrieval} & Brain Tumor \\
& Pathology & EBHI-Seg~\cite{shi2023ebhi} & Adenocarcinoma \\
& Ultrasound & TNUI-2021~\cite{zhou2022h} & Thyroid Nodule \\
& Ultrasound & BUSI~\cite{al2020dataset} & Breast Lesion \\
& Endoscopy & Colon Polyp (5 sets)~\cite{bernal2015wm,jha2019kvasir,silva2014toward,tajbakhsh2015automated,vazquez2017benchmark} & Colorectal Polyp \\
& CT & COVID-19~\cite{fan2020inf} & Lung Infection \\
& Dermoscopy & ISIC 2018~\cite{codella2019skin} & Skin Lesion \\ \midrule

\multirow{5}{*}{\begin{tabular}[c]{@{}l@{}}\textbf{Zero-Shot Phase I}\\ (Unseen Datasets,\\ Seen Modalities)\end{tabular}} 
& Ultrasound & TN3K~\cite{gong2021multi} & Thyroid Nodule \\
& Ultrasound & BUSBRA~\cite{gomez2024bus} & Breast Lesion \\
& Dermoscopy & PH2~\cite{mendoncca2015ph2} & Skin Lesion \\
& MRI (Cardiac) & ACDC~\cite{bernard2018deep} & Cardiac Structure \\
& Pathology & GlaS~\cite{sirinukunwattana2017gland} & Glandular Structure \\ \midrule

\multirow{3}{*}{\begin{tabular}[c]{@{}l@{}}\textbf{Zero-Shot Phase II}\\ (Completely Unseen \\ Modalities)\end{tabular}} 
& X-Ray & Montgomery~\cite{jaeger2014two} & Lung \\
& X-Ray & Covidquex~\cite{tahir2021covid} & Lung Infection \\
& Electron Microscopy & ISBI EM~\cite{cardona2010integrated} & Neuronal Structure \\ \bottomrule
\end{tabular}%
}
\end{table}

\subsection{In-Domain Datasets}
During the training phase, following~\cite{chang2025unified}, we utilize the same dataset composition, with 8 diverse datasets. Details of training datasets are as follows:

\noindent\textbf{AMD-SD dataset.} The AMD-SD dataset~\cite{hu2024amd} contains Optical Coherence Tomography (OCT) images specifically annotated for the segmentation of retinal lesions. Targeting Wet Age-related Macular Degeneration (AMD), this comprehensive dataset comprises 3,049 scans collected from 138 patients.

\noindent\textbf{BTD dataset.} The Brain Tumor Dataset (BTD)~\cite{cheng2015enhanced,cheng2016retrieval} consists of T1-weighted contrast-enhanced Magnetic Resonance Imaging (MRI) scans. It contains 3,064 pairs of MRI brain images annotated at the pixel level.

\noindent\textbf{EBHI-Seg dataset.} The EBHI-Seg dataset~\cite{shi2023ebhi} comprises H\&E stained histopathology images collected for the precise segmentation of adenocarcinoma and normal glandular structures. It includes 4,456 images across 6 histopathological classes.

\noindent\textbf{TNUI-2021 dataset.} The TNUI dataset~\cite{zhou2022h} provides a collection of grayscale thyroid ultrasound images expertly annotated for thyroid nodule boundaries. Comprising 1,381 meticulously delineated images, it serves as a critical benchmark.

\noindent\textbf{BUSI dataset.} The Breast Ultrasound Images (BUSI) dataset~\cite{al2020dataset} includes comprehensive ultrasound scans covering normal, benign, and malignant breast lesions. Collected from 600 female patients (aged 25 to 75), it provides 780 images ,covering 133 normal cases, 487 benign cases, and 210 malignant cases, each with corresponding ground truth.

\noindent\textbf{Colon Polyp dataset.} To ensure robustness against endoscopic device variations, following~\cite{chang2025unified}, the dataset aggregate five standard colonoscopy benchmarks, CVC-ClinicDB~\cite{bernal2015wm}, Kvasir~\cite{jha2019kvasir}, CVC-ColonDB~\cite{tajbakhsh2015automated}, ETIS-Larib~\cite{silva2014toward}, and EndoScene~\cite{vazquez2017benchmark}, into a single large-scale dataset.

\noindent\textbf{COVID-19 dataset.} This dataset~\cite{fan2020inf} contains chest CT slices meticulously annotated for lung infections. It consists of 100 axial slices from over 40 patients.

\noindent\textbf{ISIC 2018 dataset.} The ISIC 2018 dataset~\cite{codella2019skin} is a large-scale high-resolution dermoscopy image collection provided by the International Skin Imaging Collaboration. ISIC 2018 contains 2,594 dermoscopic lesion segmentation images, each with corresponding ground truth.


\subsection{Zero-shot datasets with seen modalities}
To evaluate the model's robustness and generalization ability, we freeze the network weights and directly test it on five additional unseen datasets. These datasets share the same imaging physics as the training dataset but present severe intra-domain shifts due to different clinical centers, scanning equipment, and patient demographics. Details of zero-shot inference datasets are as follows:

\noindent\textbf{TN3K dataset.} The TN3K dataset~\cite{gong2021multi} is an independent, large-scale thyroid ultrasound dataset. It comprises 3,493 representative images from 2,421 patients, carefully filtered to exclude heavy color Doppler artifacts.

\noindent\textbf{BUSBRA dataset.} The BUSBRA dataset~\cite{gomez2024bus} features breast ultrasound images collected from a different demographic. Providing anonymized scans from 1,064 patients, it includes biopsy-proven tumor cases systematically categorized with BI-RADS (categories 2-5) annotations.

\noindent\textbf{PH2 dataset.} The PH2 database~\cite{mendoncca2015ph2} contains dermoscopic images of melanocytic lesions. Acquired at Pedro Hispano Hospital, it consists of 200 high-quality images and their corresponding masks.

\noindent\textbf{ACDC dataset.} The Automated Cardiac Diagnosis Challenge (ACDC) dataset~\cite{bernard2018deep} comprises cardiac cine-MRI scans from 150 cases, targeting both end-diastolic and end-systolic frames. It aims to segment left/right ventricles and myocardium.

\noindent\textbf{GlaS dataset.} From the MICCAI 2015 Gland Segmentation Challenge~\cite{sirinukunwattana2017gland}, this dataset evaluates zero-shot glandular structure segmentation capabilities. It includes 165 images derived from 16 H\&E stained slides of T3/T4 stage colorectal adenocarcinoma.

\subsection{Zero-shot datasets with completely unseen modalities}
To evaluate the model's generalization ability to abstract universal priors, we freeze the network weights and directly test it on three additional unseen datasets with completely unseen modalities. These datasets introduce entirely different imaging principles never encountered during training, strictly stress-testing the disentangled geometric tokens' ability to map pure pixel-intensity gradients to universal geometric shapes. Details of  datasets are as follows:

\noindent\textbf{Montgomery dataset (X-Ray).} The Montgomery County dataset~\cite{jaeger2014two}, created in collaboration with the National Library of Medicine, contains 138 PA chest X-rays (80 normal, 58 abnormal/TB).

\noindent\textbf{Covidquex dataset (X-Ray).} The COVID-QU-Ex dataset~\cite{tahir2021covid} is a massive collection of 33,920 chest X-ray images, encompassing COVID-19, viral/bacterial pneumonia, and normal cases.

\noindent\textbf{ISBI EM dataset (EM).} The ISBI 2012 Challenge dataset~\cite{cardona2010integrated} focuses on segmenting highly filamentous neuronal structures from Electron Microscopy (EM) images. The ultrastructural details (such as cell membranes) present a stark morphological contrast to macroscopic lesions, serving as a critical failure-case stress test for the topological inductive bias (compactness) internalized by our geometric tokens.

\section{Experimental details and evaluation metrics}
\label{sec:extended_setup}

\noindent\textbf{Experimental details.} To ensure a rigorously fair comparison, all baseline methods and our proposed Concept-to-Pixel (C2P) framework are trained and evaluated on the exact same dataset splits across all experiments. During the training phase, to prevent overfitting and enhance model robustness, we employ a unified data augmentation pipeline across our method and all reproduced baselines. The standard augmentations include:
\begin{itemize}
    \item {Spatial Transformations:} Random horizontal and vertical flips, and random rotations within the range of $[-10^\circ, 10^\circ]$. Crucially, whenever spatial augmentations are applied to an image, the corresponding ground-truth geometric properties (\eg, bounding box, centroid, orientation) for the \texttt{[GEO]} tokens are dynamically re-computed and re-aligned on the fly to preserve strict spatial consistency.
    \item {Photometric Distortions:} Random adjustments to brightness and contrast, alongside the injection of Gaussian noise.
\end{itemize}

\noindent\textit{Note}: the only exception regarding data augmentation is nnU-Net V2~\cite{isensee2021nnu}. Since nnU-Net relies heavily on its intrinsic, highly optimized self-configuring data augmentation pipeline to achieve its performance as a specialized model, we retain its default augmentation strategy to reflect its maximum capability.

\noindent\textbf{Evaluation metrics.} We employ the widely adopted Dice Similarity Coefficient (DSC) as the primary evaluation metric. The DSC is formulated as:
\begin{equation}
    \text{Dice} = \frac{2 \times |P \cap G|}{|P| + |G|} \times 100\%
\end{equation}
where $P$ denotes the set of predicted foreground pixels, and $G$ represents the set of ground-truth foreground pixels.

\section{Offline Knowledge Pre-processing}
\label{sec:offline_preprocessing}

To ensure training efficiency and to provide stable, deterministic targets for the explicit decoupling mechanism in our C2P, we pre-compute both the geometric properties and the semantic text embeddings offline for all training samples.

\subsection{Offline Geometric Property Extraction}
The geometric properties serve as explicit regression targets for the \texttt{[GEO]}. To avoid the significant computational overhead of processing high-resolution binary masks during the training loop, we systematically extract these properties offline. 

Specifically, all ground-truth masks are first uniformly resized to the training resolution ($384 \times 384$) using nearest-neighbor interpolation to rigorously preserve their discrete binary nature. Let $\mathcal{M} \in \{0, 1\}^{H \times W}$ denote the binary mask, and $\Omega = \{(x, y) \mid \mathcal{M}(x,y) = 1\}$ be the set of foreground pixels with absolute pixel area $A = |\Omega|$. We select nine geometric properties, which are mathematically defined as follows:

\begin{itemize}
    \item \textbf{Area ($\hat{A}$):} The foreground area is computed as: $\hat{A} = \frac{A}{H \times W}$.
    
    \item \textbf{Centroid ($\hat{C}_x, \hat{C}_y$):} The center of foreground mass is computed as:
    \begin{equation}
        \hat{C}_x = \frac{1}{W \cdot A} \sum_{(x,y) \in \Omega} x, \quad \hat{C}_y = \frac{1}{H \cdot A} \sum_{(x,y) \in \Omega} y.
    \end{equation}
    
    \item \textbf{Bounding Box:} The bounding box 
    of foreground is computed as:
    \begin{equation}
        \left[ \frac{\min x}{W}, \frac{\min y}{H}, \frac{\max x}{W}, \frac{\max y}{H} \right] \quad \text{for } (x,y) \in \Omega.
    \end{equation}
    
    \item \textbf{Aspect Ratio ($\hat{AR}$):} Let $w_{box}$ and $h_{box}$ be the absolute width and height of the bounding box. The ratio is computed as: $\hat{AR} = \frac{1}{10} \min\left(\frac{w_{box}}{h_{box}}, 10\right)$.
    
    \item \textbf{Perimeter ($\hat{P}$):} The foreground perimeter is computed as: $\hat{P} = \frac{P}{2(H + W)}$.
    
    \item \textbf{Compactness ($Comp$):} The shape circularity of foreground is computed as:
    \begin{equation}
        Comp = \min\left(\frac{4\pi A}{P^2}, 1.0\right).
    \end{equation}
    
    \item \textbf{Solidity ($Sol$):} The solidity of foreground is computed as: $Sol = \frac{A}{A_{convex}}$.
    
    \item \textbf{Eccentricity ($Ecc$):} The eccentricity of foreground is computed as: $Ecc = \frac{c}{a}$, where $c$ is the focal distance and $a$ is the major axis length.
    
    \item \textbf{Orientation ($\hat{\theta}$):} Let $\theta \in [-\frac{\pi}{2}, \frac{\pi}{2}]$ be the angle between the major axis of the fitted ellipse and the x-axis. It is linearly mapped to $[0, 1]$: $\hat{\theta} = \frac{\theta + \pi/2}{\pi}$.
\end{itemize}

In edge cases where a mask is completely empty (\ie, $A = 0$), a predefined set of default zero and center values (\eg, $\hat{C}_x=0.5, \hat{C}_y=0.5, \hat{AR}=1.0$) is assigned to ensure numerical stability.

\subsection{Offline Semantic Knowledge Generation}
To distill expert-level clinical concepts into the \texttt{[SEM]} tokens, we utilize a Multimodal Large Language Model (MLLM) to generate structured medical reports and subsequently encode them into dense embeddings.

\noindent\textbf{Visual Triplet Prompting.} Modern MLLMs often struggle to accurately localize tiny medical lesions within complex anatomical backgrounds when only provided with raw images. To address this, we design a ``Visual Triplet'' prompting strategy. For each sample, we feed the MLLM with three explicit visual cues:
\begin{enumerate}
    \item \textbf{Image 1 (raw):} The original, un-annotated image for detailed internal texture and pattern analysis.
    \item \textbf{Image 2 (annotated):} The original image overlaid with a clearly dilated contour (derived from the ground-truth mask) to explicitly guide the MLLM's attention to the Region of Interest (ROI).
    \item \textbf{Image 3 (mask):} The binary mask to unambiguously present the exact geometric shape and boundary contour.
\end{enumerate}

\noindent\textbf{Structured Clinical Reasoning.} Alongside the visual triplet, we employ a highly structured text prompt. We assign the MLLM a specific expert role and instruct it to perform a detailed analysis across multiple diagnostic dimensions (e.g., morphology, margins, internal texture, and surrounding interaction). Finally, the model is constrained to output a strict JSON object covering 9 explicit fields. A complete example of our prompt template, instantiated for the Breast Ultrasound (BUSI) dataset, is illustrated in Fig.~\ref{fig:prompt}. Consequently, upon processing these multimodal inputs, the MLLM generates a standardized diagnostic report, as presented in Fig.~\ref{fig:response}. This highly structured JSON format ensures that the extracted clinical semantics are consistent, comprehensive, and readily encodable into the dense \texttt{[SEM]} embeddings.

\noindent\textbf{Dataset-Specific Prompt Adaptations.} 
While the base structure of the prompt (i.e., Visual Analysis Strategy, Input Description, and JSON Output Requirement) remains identical across all modalities to ensure standardized output formats, we meticulously tailor the \textbf{Role}, \textbf{Specific Analysis Tasks}, \textbf{Image Mode}, and \textbf{Diagnostic Context} to inject domain-specific medical priors for the remaining 7 datasets. These specific variations are detailed below:

\begin{figure*}[htpb]
    \centering
    \includegraphics[width=\textwidth, trim=0.1cm 0.1cm 0.1cm 0.1cm, clip]{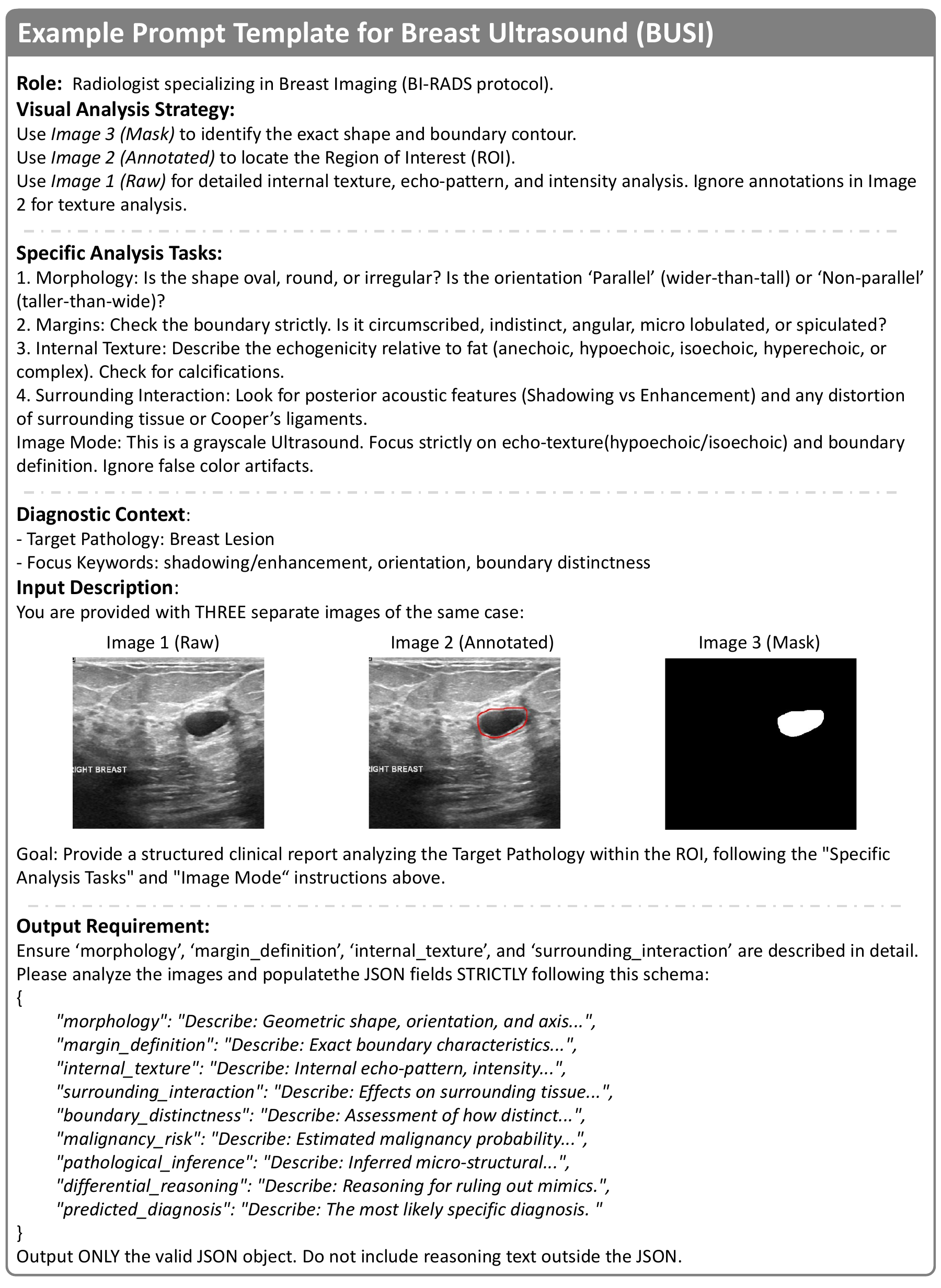} 
    \caption{\textbf{Example Prompt Template for Breast Ultrasound (BUSI) used for Offline Semantic Knowledge Generation.}}
    \label{fig:prompt}
\end{figure*}

\begin{figure*}[htpb]
    \centering
    \includegraphics[width=\textwidth, trim=0.1cm 0.1cm 0.1cm 0.1cm, clip]{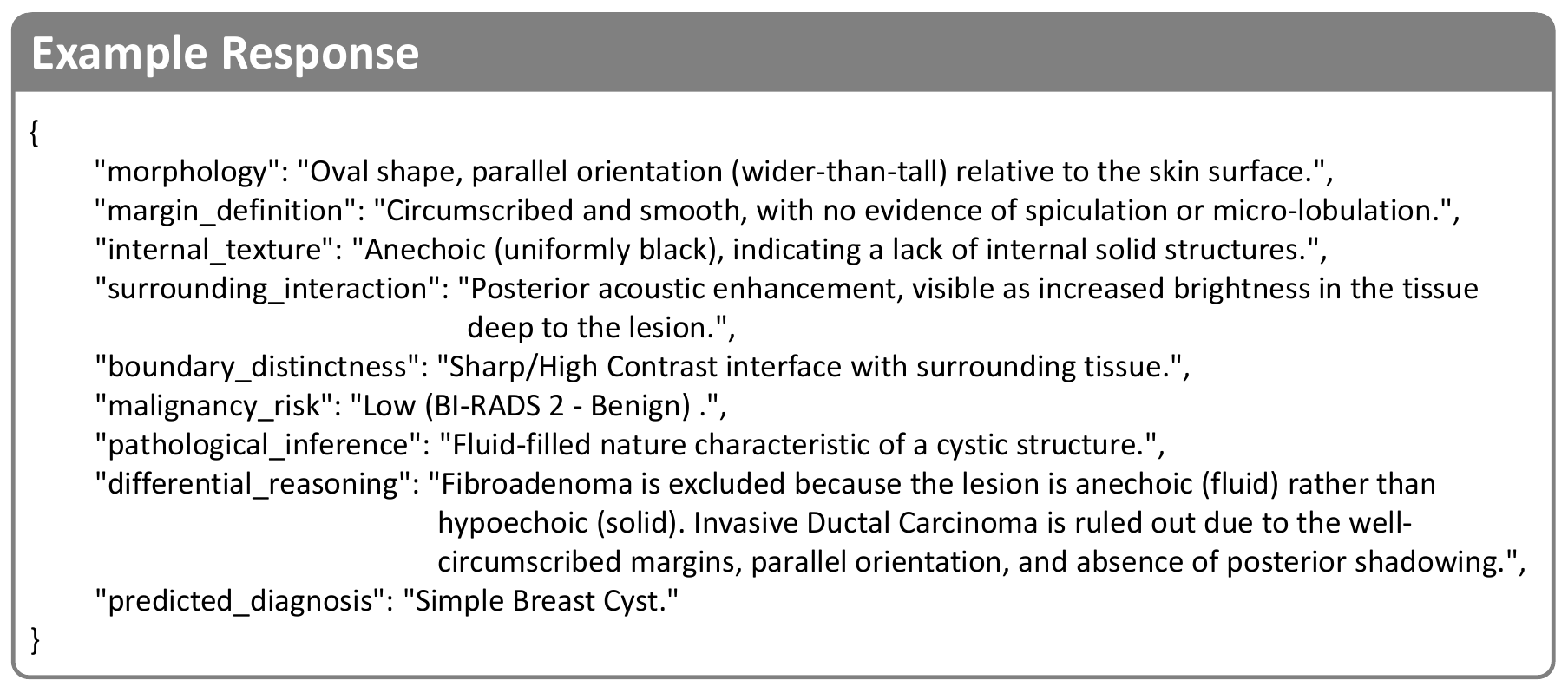} 
    \caption{\textbf{Example of the structured JSON response generated by the MLLM.} Corresponding to the input prompt in Fig.~\ref{fig:prompt}, the model successfully outputs a comprehensive clinical analysis strictly following the predefined 9-dimension schema.}
    \label{fig:response}
\end{figure*}

\begin{itemize}
    \item \textbf{Wet AMD (AMD-SD dataset) | Modality: OCT}
    \begin{itemize}
        \item \textbf{Role:} Expert Ophthalmologist specializing in Retinal OCT analysis.
        \item \textbf{Specific Analysis Tasks:} \textbf{(1) Morphology}: Describe the shape of fluid accumulation (cystoid, dome-shaped, or flat). \textbf{(2) Margins}: Are the boundaries of the fluid/lesion distinct? \textbf{(3) Internal Texture}: Distinguish between Intraretinal Fluid (IRF), Subretinal Fluid (SRF), and PED. Look for Hyper-reflective foci. \textbf{(4) Surrounding Interaction}: Analyze the disruption of retinal layers (ELM, IS/OS junction) and RPE integrity.
        \item \textbf{Image Mode:} Grayscale OCT scan. Focus strictly on grayscale intensity gradients (hyper-reflective vs hypo-reflective) and layer disruption.
        \item \textbf{Context:} Target Pathology: Retinal Lesion (Wet AMD). Keywords: layer disruption (ILM/RPE), fluid accumulation, shadowing effect, signal intensity.
    \end{itemize}

    \item \textbf{Brain Tumor (BTD dataset) | Modality: MRI}
    \begin{itemize}
        \item \textbf{Role:} Senior Neuroradiologist specializing in Brain Tumor MRI (T1-weighted).
        \item \textbf{Specific Analysis Tasks:} \textbf{(1) Morphology}: Describe the overall shape (e.g., ring-like, irregular mass). \textbf{(2) Margins}: Is the boundary well-defined or infiltrating into the white matter? \textbf{(3) Internal Texture}: Analyze signal heterogeneity. Identify the 'Enhancing Rim' vs 'Necrotic Core'. \textbf{(4) Surrounding Interaction}: Describe the 'Mass Effect' (midline shift, ventricular compression) and peritumoral edema.
        \item \textbf{Context:} Target Pathology: Brain Tumor. Keywords: enhancing rim vs necrotic core, mass effect, boundary definition.
    \end{itemize}

    \item \textbf{Adenocarcinoma (EBHI-Seg dataset) | Modality: Histopathology}
    \begin{itemize}
        \item \textbf{Role:} Expert Pathologist specializing in Colorectal Histopathology (H\&E stain).
        \item \textbf{Specific Analysis Tasks:} \textbf{(1) Morphology}: Analyze the glandular architecture. Are glands well-formed, cribriform, fused, or single cells? \textbf{(2) Margins}: Characterize the invasion front ('Pushing' or 'Infiltrative'). \textbf{(3) Internal Texture}: Evaluate cellular atypia. Look for high N/C ratio, hyperchromasia, and pleomorphism. \textbf{(4) Surrounding Interaction}: Look for 'Desmoplastic reaction' (stromal fibrosis) or lymphocytic infiltration.
        \item \textbf{Image Mode:} Stained color image. Pay strict attention to color variegation (e.g., purple nuclear crowding vs. pink cytoplasm) as an indicator of malignancy.
        \item \textbf{Context:} Target Pathology: Adenocarcinoma. Keywords: glandular differentiation, N/C ratio, cell crowding, infiltrative margin.
    \end{itemize}

    \item \textbf{Thyroid Nodule (TNUI dataset) | Modality: Ultrasound}
    \begin{itemize}
        \item \textbf{Role:} Radiologist specializing in Thyroid Imaging (ACR TI-RADS protocol).
        \item \textbf{Specific Analysis Tasks:} \textbf{(1) Morphology}: Evaluate the aspect ratio ('Taller-than-wide' or 'Wider-than-tall'). \textbf{(2) Margins}: Is the margin smooth, ill-defined, micro-lobulated, or spiculated? \textbf{(3) Internal Texture}: Describe the echogenicity. Look specifically for 'Microcalcifications'. \textbf{(4) Surrounding Interaction}: Check for a 'Halo sign' or extrathyroidal extension.
        \item \textbf{Context:} Target Pathology: Thyroid Nodule. Keywords: aspect ratio, halo sign, microcalcification, echogenicity.
    \end{itemize}

    \item \textbf{Colon Polyp (Polyp dataset) | Modality: Endoscopy}
    \begin{itemize}
        \item \textbf{Role:} Gastroenterologist specializing in Colonoscopy and NBI imaging.
        \item \textbf{Specific Analysis Tasks:} \textbf{(1) Morphology}: Classify based on Paris Classification (Pedunculated, Sessile, Flat). \textbf{(2) Margins}: Analyze the demarcation line between the lesion and normal mucosa. \textbf{(3) Internal Texture}: Describe the surface 'Pit Pattern' (Kudo classification) and vascular intensity. \textbf{(4) Surrounding Interaction}: Look for convergence of mucosal folds or mucosal disruption.
        \item \textbf{Image Mode:} Color endoscopic image. Pay strict attention to color variations (redness, fading) and vascular color patterns.
        \item \textbf{Context:} Target Pathology: Colorectal Polyp. Keywords: surface vessel pattern, pit pattern, stalk presence.
    \end{itemize}

    \item \textbf{Lung Infection (COVID-19 dataset) | Modality: CT}
    \begin{itemize}
        \item \textbf{Role:} Thoracic Radiologist specializing in Pulmonary CT.
        \item \textbf{Specific Analysis Tasks:} \textbf{(1) Morphology}: Describe the distribution (Peripheral, Basal, or Diffuse). \textbf{(2) Margins}: Is the opacity 'Ill-defined' or 'Sharp'? \textbf{(3) Internal Texture}: Distinguish between 'Ground-Glass Opacity' (GGO) and 'Consolidation'. Look for 'Crazy-paving'. \textbf{(4) Surrounding Interaction}: Check for 'Air bronchograms', vascular thickening, or pleural traction.
        \item \textbf{Image Mode:} Grayscale CT scan. Focus strictly on Hounsfield Unit density representation.
        \item \textbf{Context:} Target Pathology: Lung Infection. Keywords: Ground-Glass Opacity (GGO), consolidation, pleural delimitation.
    \end{itemize}

    \item \textbf{Skin Lesion (ISIC 2018 dataset) | Modality: Dermoscopy}
    \begin{itemize}
        \item \textbf{Role:} Dermatologist specializing in Dermoscopy.
        \item \textbf{Specific Analysis Tasks:} \textbf{(1) Morphology}: Evaluate Asymmetry (1 or 2 axes). \textbf{(2) Margins}: Is the border regular or irregular (map-like, scalloped)? Is there an abrupt cutoff? \textbf{(3) Internal Texture}: Look for specific structures: 'Pigment Network', 'Dots/Globules', 'Streaks', or 'Blue-white veil'. \textbf{(4) Surrounding Interaction}: Look for regression structures or peripheral inflammation.
        \item \textbf{Image Mode:} High-resolution color image. Describe color variegation (e.g., mixture of dark brown, slate blue, white, red).
        \item \textbf{Context:} Target Pathology: Skin Lesion. Keywords: pigment network, border abruptness, textural homogeneity.
    \end{itemize}
\end{itemize}

\noindent\textbf{Structured Clinical Reasoning.} Alongside the visual triplet, we employ a highly structured text prompt. We assign the MLLM a specific expert role and instruct it to output a strict JSON object covering 9 explicit diagnostic dimensions. To provide a concrete illustration of our prompt engineering, the complete prompt template instantiated for the Breast Ultrasound (BUSI) dataset is presented below:

\begin{itemize}

\item ``\textit{The geometric shape and orientation of the lesion is} \texttt{\{\texttt{Morphology}\}}''.

\item ``\textit{The boundaries and margins of the lesion are} \texttt{\{\texttt{Margin Definition}\}}''.

\item ``\textit{The internal echo-texture and composition is} \texttt{\{\texttt{Internal Texture}\}}''.

\item ``\textit{The interaction with surrounding tissue shows} \texttt{\{\texttt{Surrounding Interaction}\}}''.

\item ``\textit{The visual distinctness and contrast of the lesion edge is} \texttt{\{\texttt{Boundary Distinctness}\}}''.

\item ``\textit{The estimated clinical malignancy risk assessment is} \texttt{\{\texttt{Malignancy Risk}\}}''.

\item ``\textit{The inferred micro-structural tissue characteristics are} \texttt{\{\texttt{Pathological Inference}\}}''.

\item ``\textit{The clinical reasoning for distinguishing this from other mimics is} \texttt{\{\texttt{Differential Reasoning}\}}''.

\item ``\textit{The most likely specific diagnosis is} \texttt{\{\texttt{Predicted Diagnosis}\}}''.

\end{itemize}

Following this template injection, the 9 complete descriptive sentences are processed independently by a pre-trained domain-specific language encoder (\texttt{PubMedBERT~\cite{gu2021domain}}). Instead of using individual word embeddings, we extract the final hidden state corresponding to the \texttt{[CLS]} token, which serves as the global semantic representation of the entire description. This yields an $N_{sem} \times 768$ dense embedding matrix (where $N_{sem}=9$) for each image, which is saved as an \texttt{.npy} file. During the subsequent network training, these frozen embeddings serve as the explicit contrastive alignment targets for our learnable semantic tokens.

\section{Analysis}

\subsection{Ablation of Different MLLMs}
To evaluate the impact of different foundation models on semantic knowledge generation, we conduct an ablation study by replacing the core MLLM generator. As reported in Table~\ref{tab:ablation_mllms}, we benchmarked five different MLLMs: Doubao-Seed-1.6-flash, Grok 4 Fast, Gemini 3.0 Pro, GPT-4o mini, and Qwen3-VL-Max. During this experiment, the visual triplet inputs and the structured clinical reasoning prompts remained strictly identical across all models.

Overall, all evaluated MLLMs demonstrate strong and stable performance, with average Dice scores consistently exceeding 87.5\%. This indicates that our proposed visual triplet prompting strategy and structured JSON schema are highly effective and model-agnostic. They successfully elicit valuable clinical priors and robust semantic representations regardless of the specific underlying MLLM.

However, fine-grained modality performance reveals distinct model characteristics. For instance, Doubao excels in Endoscopy (Polyp, 92.72\%) and Breast Ultrasound (81.12\%), while Gemini and GPT-4o mini achieve the highest accuracy in Dermoscopy (ISIC2018, 90.40\%) and MRI (BTD, 85.87\%). Notably, Qwen3-VL-Max achieves the highest overall average Dice score of 87.79\%. It demonstrates exceptional robustness and precise visual grounding in challenging modalities such as Chest CT (COVID, 83.49\%) and Thyroid Ultrasound (TNUI, 88.52\%). Given its superior comprehensive performance and balanced cross-modality reasoning capabilities, we utilize Qwen3-VL-Max as our default MLLM generator.

\begin{table*}[tb]
  \caption{Ablation study on the choice of different Multimodal Large Language Models (MLLMs) for semantic knowledge generation (Dice \%). The best results are highlighted in \textbf{bold}.}
  \label{tab:ablation_mllms}
  \centering
  \setlength{\tabcolsep}{3pt} 
  \resizebox{\textwidth}{!}{%
  \begin{tabular}{@{}lccccccccc@{}}
    \toprule
    \multirow{3}{*}{MLLM Generator} & \multicolumn{8}{c}{Modalities} & \multirow{3}{*}{Avg} \\
    \cmidrule(lr){2-9} 
     & Pathology & \multicolumn{2}{c}{Ultrasound} & Endoscopy & CT & Dermoscopy & OCT & MRI & \\

     & \makebox[1.4cm][c]{EBHI} & \makebox[1.4cm][c]{TNUI} & \makebox[1.4cm][c]{Breast} & \makebox[1.4cm][c]{Polyp} & \makebox[1.4cm][c]{COVID} & \makebox[1.4cm][c]{ISIC2018} & \makebox[1.4cm][c]{AMDSD} & \makebox[1.4cm][c]{BTD} &  \\
    \midrule
    
    Doubao-Seed-1.6-flash & 95.00 & 88.14 & \textbf{81.12} & \textbf{92.72} & 82.64 & 90.09 & 85.69 & 85.13 & 87.57 \\
    Grok 4 Fast           & 94.94 & 88.35 & 80.90 & 92.66 & 81.99 & 90.35 & \textbf{86.10} & 85.34 & 87.58 \\
    Gemini 3.0 Pro        & \textbf{95.08} & 88.23 & 81.02 & 92.33 & 82.68 & \textbf{90.40} & 85.96 & \textbf{85.87} & 87.70 \\
    GPT-4o mini           & 95.02 & 88.45 & 80.78 & 92.48 & 82.81 & \textbf{90.40} & 86.05 & \textbf{85.87} & 87.73 \\
    Qwen3-VL-Max          & 95.04 & \textbf{88.52} & 80.90 & 92.65 & \textbf{83.49} & 90.25 & 86.07 & 85.43 & \textbf{87.79} \\
    
    \bottomrule
  \end{tabular}%
  }
\end{table*}

\subsection{Qualitative Analysis on Zero-Shot Generalization}
\label{sec:qualitative_unseen}

\begin{figure}[htpb]
  \centering
  \includegraphics[width=\textwidth]{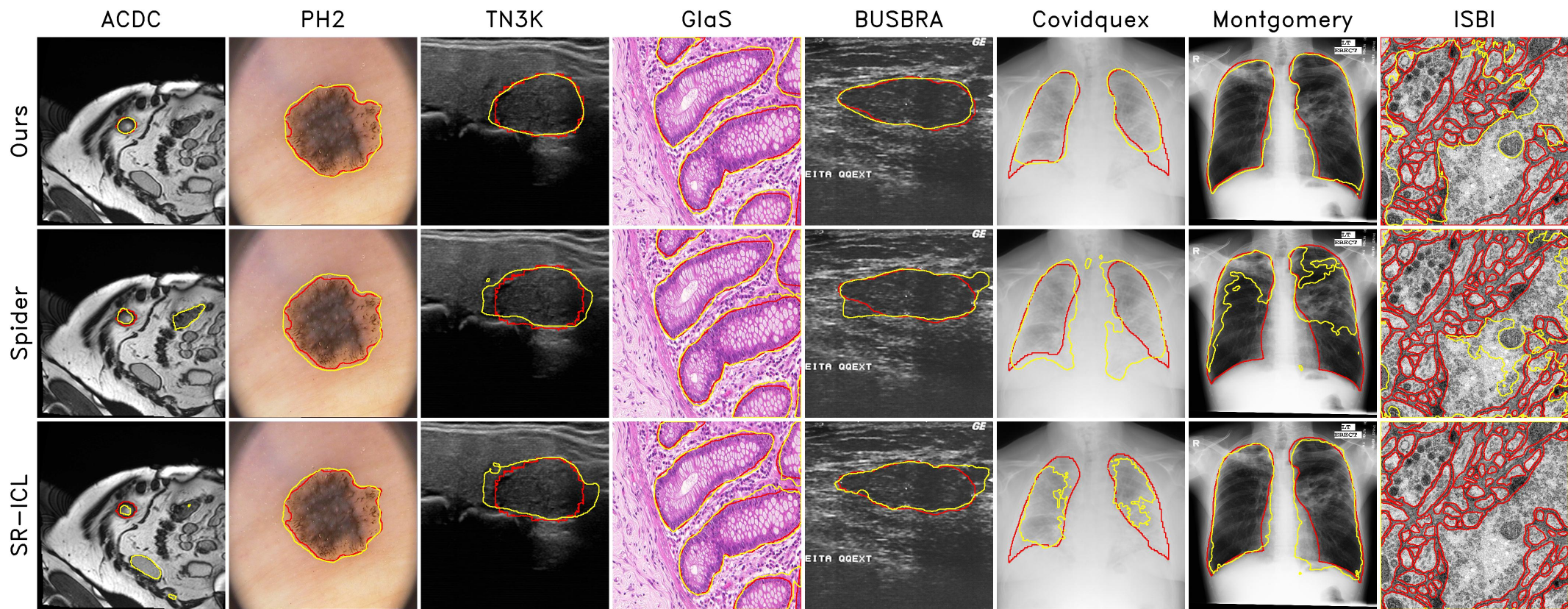} 
  \caption{\textbf{Qualitative zero-shot generalization results.} From left to right: ACDC, PH2, TN3K, GlaS, BUSBRA, Covidquex, Montgomery, and ISBI EM. \textcolor{red}{Red} indicates Ground Truth, and \textcolor{yellow}{yellow} denotes predictions. C2P (top) shows superior robustness against domain shifts and remarkable generalization to unseen macroscopic modalities (\eg, X-Ray). Conversely, its learned compact geometric prior causes expected under-segmentation on the filamentous structures in ISBI EM.}
  \label{fig:qualitative_unseen}
\end{figure}

To further illustrate the zero-shot generalization capabilities of C2P, we provide visual comparisons against Spider and SR-ICL across 8 unseen datasets in \cref{fig:qualitative_unseen}.

\noindent\textbf{Robustness to Intra-Domain Shifts.} On unseen datasets of seen modalities (\eg, ACDC, BUSBRA), baseline methods struggle with domain shifts, producing fragmented boundaries or severe false positives. In contrast, C2P generates precise and geometrically plausible masks. This demonstrates that the frozen \texttt{[SEM]} robustly adapts to novel intensity distributions, while \texttt{[GEO]} strictly regularizes the object shape.

\noindent\textbf{Generalization to Unseen Macroscopic Modalities.} The contrast is starkest on entirely unseen modalities like X-Ray (Covidquex, Montgomery). In-context learning models (Spider, SR-ICL) fail catastrophically here, as their reliance on pixel-level matching collapses across massive cross-modal gaps. By relying on abstracted concepts rather than raw pixels, C2P successfully delineates lung fields with highly accurate and smooth contours.

\noindent\textbf{Topological Limitation (Failure Case).} Conversely, C2P predictably under-segments the highly filamentous neuronal membranes in the ISBI EM dataset. Since our \texttt{[GEO]} was trained exclusively on macroscopic, compact lesions, it enforces a strict "compactness" prior that actively suppresses elongated predictions. This transparently visualizes the trade-off of explicit geometric reasoning, which can be mitigated in the future by expanding the training mixture.

\subsection{Visualization of Attention Map}

To intuitively understand the internal mechanisms of our proposed framework and verify whether it successfully learns universal shape priors, we visualize the attention maps generated by the geometric tokens. These maps highlight the regions where the model focuses its attention when predicting the target structures.

\begin{figure*}[htpb]
    \centering
    \includegraphics[width=\textwidth]{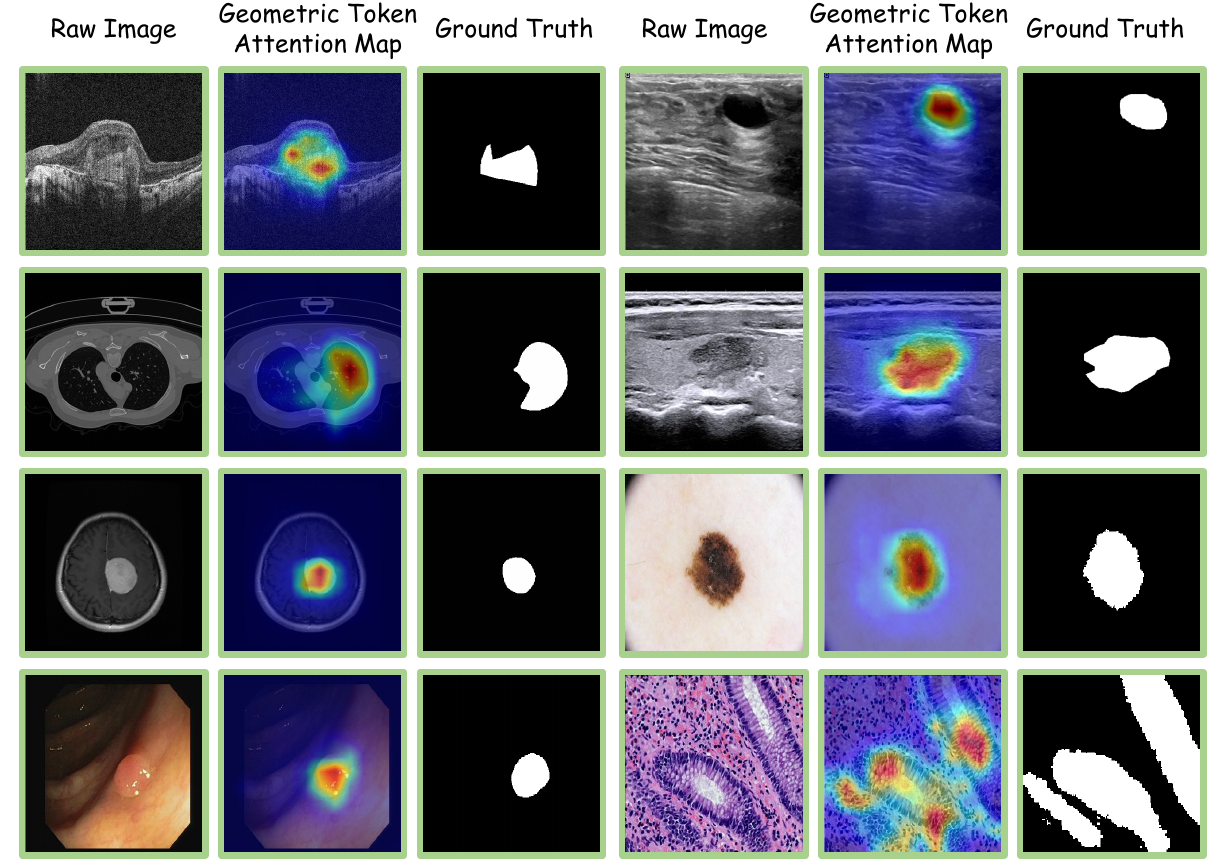} 
    \caption{Visualization of the Geometric Token Attention Maps on the in-domain validation set. The high-response areas (red/yellow) demonstrate a strict spatial alignment with the corresponding ground truth masks across highly diverse seen modalities (e.g., OCT, CT, MRI, Ultrasound, Pathology), proving the model's accurate semantic-to-visual grounding capability.}
    \label{fig:attn_indomain}
\end{figure*}

\textbf{In-Domain Localization.} As illustrated in Fig.~\ref{fig:attn_indomain}, we first evaluate the attention maps on the validation sets of the seen domains. Across highly diverse modalities---including OCT, CT, MRI, Endoscopy, Ultrasound, Dermoscopy, and Pathology---the high-response regions (warm colors) of the geometric tokens exhibit a remarkably precise alignment with the ground truth masks. This strong semantic-to-visual grounding demonstrates that the model effectively translates the structured text prompts into accurate spatial localizations, successfully isolating the target lesions from complex anatomical backgrounds.

\begin{figure*}[htpb]
    \centering
    \includegraphics[width=\textwidth]{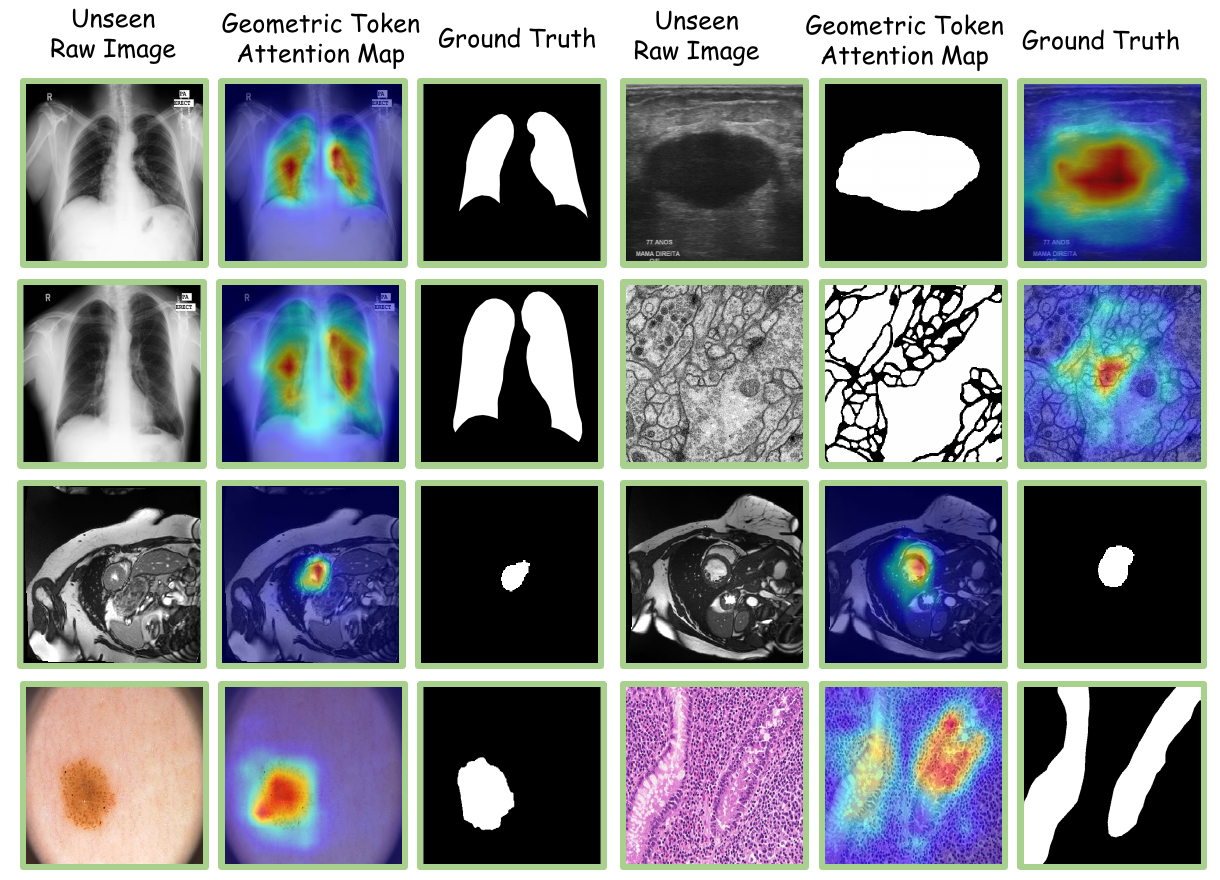} 
    \caption{Visualization of the Geometric Token Attention Maps under zero-shot generalization settings. The model successfully localizes target anatomical structures on completely unseen datasets (e.g., Cardiac MRI, unseen Dermoscopy) and entirely novel macroscopic/microscopic modalities (e.g., Chest X-Ray, Electron Microscopy). This highlights the framework's strong capability to disentangle universal geometric priors from domain-specific imaging physics.}
    \label{fig:attn_zeroshot}
\end{figure*}

\textbf{Zero-Shot Generalization.} The most compelling evidence of our model's representation power lies in its zero-shot generalization to unseen data, as shown in Fig.~\ref{fig:attn_zeroshot}. We visualize the attention maps on both unseen datasets with severe domain shifts (e.g., Cardiac MRI, PH2 Dermoscopy) and completely novel modalities never encountered during training (e.g., Chest X-Ray for lung fields, Electron Microscopy for cellular structures). Despite the drastic changes in imaging physics and pixel-intensity distributions, the geometric tokens consistently and accurately highlight the unseen target structures. This visually confirms that our framework does not merely memorize domain-specific textures; rather, it successfully abstracts and disentangles universal geometric and topological priors, enabling robust cross-modality adaptation.

